\documentclass[conference]{IEEEtran}
\usepackage{times}

\usepackage[numbers]{natbib}
\usepackage{multicol}
\usepackage{amsfonts,amssymb,amsbsy,amsmath,amsthm}
\usepackage[bookmarks=true]{hyperref}
\usepackage{moreverb,url,subcaption}
\usepackage{caption}
\usepackage{graphicx}
\usepackage{multirow}
\usepackage{algorithm}
\usepackage{algpseudocode}
\usepackage{float}
\usepackage{booktabs}
\usepackage{stfloats}
\usepackage{array}
\usepackage{dsfont}
\usepackage{color}
\newcommand{\oura}{\textsc{HAMnet}}
\newcommand{\ourr}{\textsc{UniCORN}}

\begin{document}

\title{Hierarchical and Modular Network on Non-prehensile Manipulation in General Environments}

\author{\authorblockN{Yoonyoung Cho$^*$,
Junhyek Han$^*$,
Jisu Han,
Beomjoon Kim}
\authorblockA{Korea Advanced Institute of Science and Technology}
\vspace{-3em}
}

\twocolumn[{
\renewcommand\twocolumn[1][]{#1}
\maketitle
\begin{center}
    \includegraphics[width=0.9\textwidth]{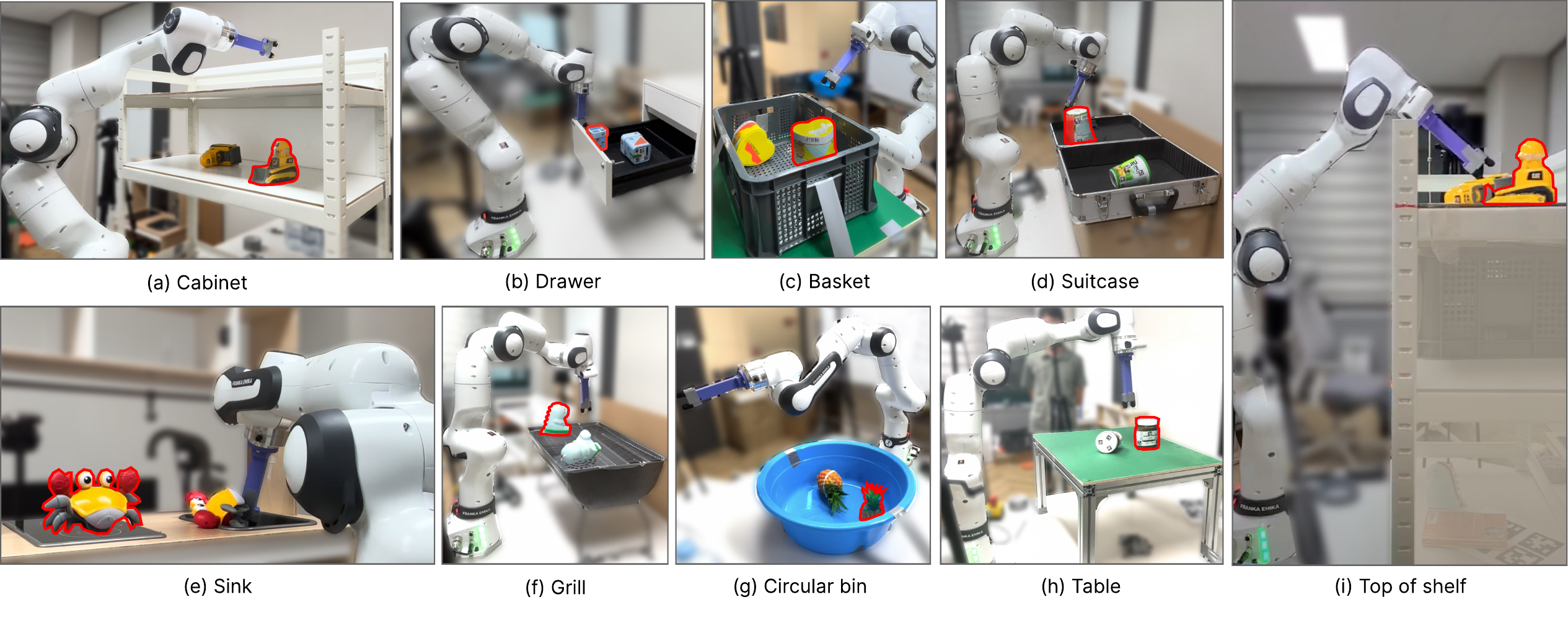}
    \captionof{figure}{\label{fig:motivation}
    Illustration of our real-world domains. Our approach generalizes to diverse and unseen objects and environments. The object outlined in red represents the goal pose. Note how the environment geometry limits object and robot motions: in (a), the robot must reach and maneuver the object while avoiding the ceiling; in (b) and (c), the robot must circumvent walls to access the object; in (d), the robot must move the object over a barrier; in (e), the robot must first traverse the sink to contact under the crab, lift it across the sink wall, then reorient the crab to face forward; and in (i), the robot must consider its kinematics while manipulating the object on a high shelf. Each object and its pose are chosen so that it cannot be simply pick-and-placed.
    }
\end{center}
}
]

\begin{abstract}
For robots to operate in general environments like households, they must be able to perform non-prehensile manipulation actions such as toppling and rolling to manipulate ungraspable objects.  However, prior works on non-prehensile manipulation cannot yet generalize across environments with diverse geometries. The main challenge lies in adapting to varying environmental constraints: within a cabinet, the robot must avoid walls and ceilings; to lift objects to the top of a step, the robot must account for the step's pose and extent. While deep reinforcement learning (RL) has demonstrated impressive success in non-prehensile manipulation, accounting for such variability presents a challenge for the generalist policy, as it must learn diverse strategies for each new combination of constraints. 
To address this, we propose a modular architecture that uses different combinations of reusable modules based on task requirements. To capture the geometric variability in environments, we extend the contact-based object representation from CORN~\cite{cho2024corn} to environment geometries, and propose a procedural algorithm for generating diverse environments to train our agent.  Taken together, the resulting policy can zero-shot transfer to novel real-world environments despite training entirely within a simulator. We additionally release a simulation-based benchmark featuring nine digital twins of real-world scenes with 353 objects to facilitate non-prehensile manipulation research in realistic domains. Code, videos, and simulation benchmarks are available on the \href{https://unicorn-hamnet.github.io/}{project website}.

\end{abstract}
\renewcommand{\thefootnote}{\fnsymbol{footnote}}
\footnotetext[1]{Equal Contribution.}
\renewcommand*{\thefootnote}{\arabic{footnote}}
\IEEEpeerreviewmaketitle

\section{Introduction}

Despite recent advances in robot manipulation, the practical deployment of robots in everyday environments like households remains challenging. One key reason is the robot's inability to manipulate ungraspable objects.
While much of prior work on manipulation centers around \textit{prehensile} manipulation~\cite{mahler2017dexnet, grasptransformer2022wang, anygrasp2023fang, wan2023unidexgrasp++}, such approaches
fall short in unstructured environments where objects are often ungraspable due to their geometry and the surrounding scene. To overcome this, robots must embrace non-prehensile manipulation, such as pushing, toppling, and rolling~\cite{lynchcontact,lynch1999np,cho2024corn}.

Recently, reinforcement learning (RL)-based approaches have achieved several successes in non-prehensile manipulation~\cite{kim2023crm,zhou2023hacman,cho2024corn,wu2024retarget}. However, these works have been limited to fixed objects in fixed scenes~\cite{kim2023crm}, general objects on flat tabletops~\cite{zhou2023hacman,cho2024corn}, or minor variations in objects and environments~\cite{wu2024retarget}.  As such, no prior work addresses non-prehensile manipulation for novel objects \textit{and} environments with arbitrary geometries as in Figure~\ref{fig:motivation}.  Based on this observation, our objective is to extend RL to enable non-prehensile manipulation in such diverse setups.

The key challenge here lies in training a policy that can adapt to the constraints imposed by the given scene. For instance, consider the scenarios in Figure~\ref{fig:motivation}: for each domain, the robot is presented with a unique set of constraints, which may also evolve during an episode (Figure~\ref{fig:motivation}e). This requires the policy to not only model multiple distinct behaviors, but also rapidly switch between behaviors in response to minor changes in state. For example, in the scenario depicted in Figure~\ref{fig:motivation}e, as soon as the toy crab is positioned above the sink, the robot must quickly switch from a \emph{lifting} skill to a \emph{translation} skill. However, standard networks struggle to learn such high-frequency functions, a phenomenon known as \emph{spectral bias}~\cite{tancik2020fourier,basri2020bias,rahaman19bias}.

\begin{figure}[t]
    \centering
    \includegraphics[width=0.9\linewidth]{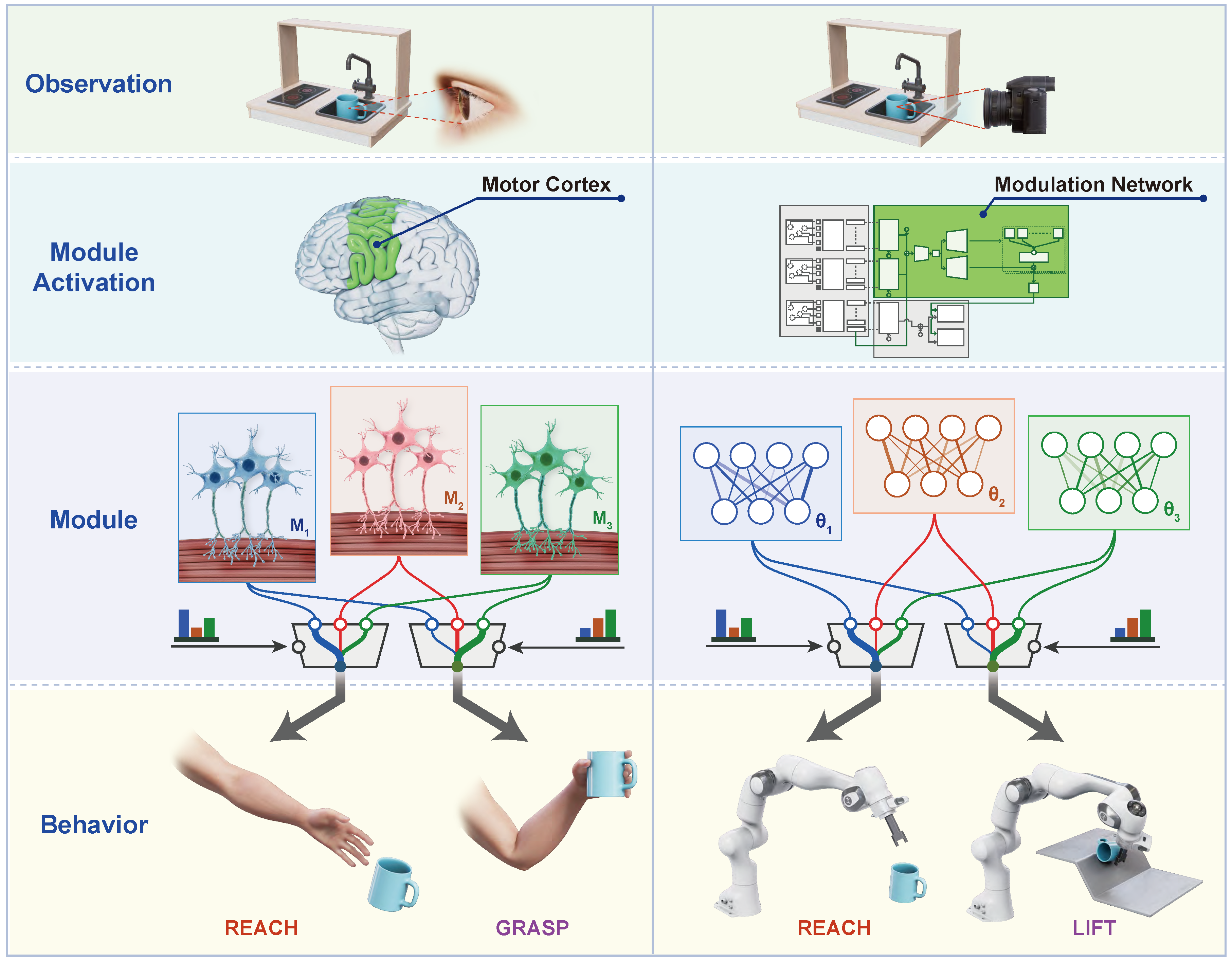}
    \caption{Illustration of computational structure for biological motor control (left) and our architecture (right).
    Each row compares analogous components in (1) acquiring sensory observations, (2) determining module activations, (3) modules representing groups of co-activated neurons, and (4) composing the modules to construct specific behaviors. The bar graphs denote the activation weight of each module.  
    }
    \label{fig:bio-figure}
    \vspace{-2ex}
\end{figure}

Human brains, on the other hand, are extremely adaptive, and their computational structure for actions differs significantly from that of standard artificial neural networks. The computation of our brain is organized modularly~\cite{clune2013modular}, where a motor cortex orchestrates neural activities at the level of \emph{motor modules}, a group of co-activating motor neurons\footnote{Neurons connected to muscle fibers that trigger muscular contractions.}~\cite{Tresch1999spinal,Overduin2012monkey}, to produce an action based on the current task~\cite{Overduin2015RepresentationOM,ting2007neuromechanics,kantak2011rewiring}. By invoking different sets of modules in response to the current context, the motor system produces disjoint behaviors such as reaching or grasping without interference~\cite{ellefsen2015forget}. This is illustrated in Figure~\ref{fig:bio-figure}, left.

Inspired by this, we propose a modular and hierarchical policy architecture (Figure~\ref{fig:bio-figure}, right). In our architecture, the \emph{modulation network} assumes the role of the motor cortex which determines the activations of modules, each representing a group of co-activated network parameters. The weighted combinations of these modules are then used to define the parameters of a \emph{base network}, which has a fixed architecture but the parameters are determined online by the modulation network based on the current context, such as the environmental constraints. Unlike standard neural networks, this computational structure enables a single base network to model multiple functions, allowing the policy to produce distinct behaviors in response to even subtle changes in context.

\begin{figure}[t]
    \centering
    \includegraphics[width=0.9\linewidth]{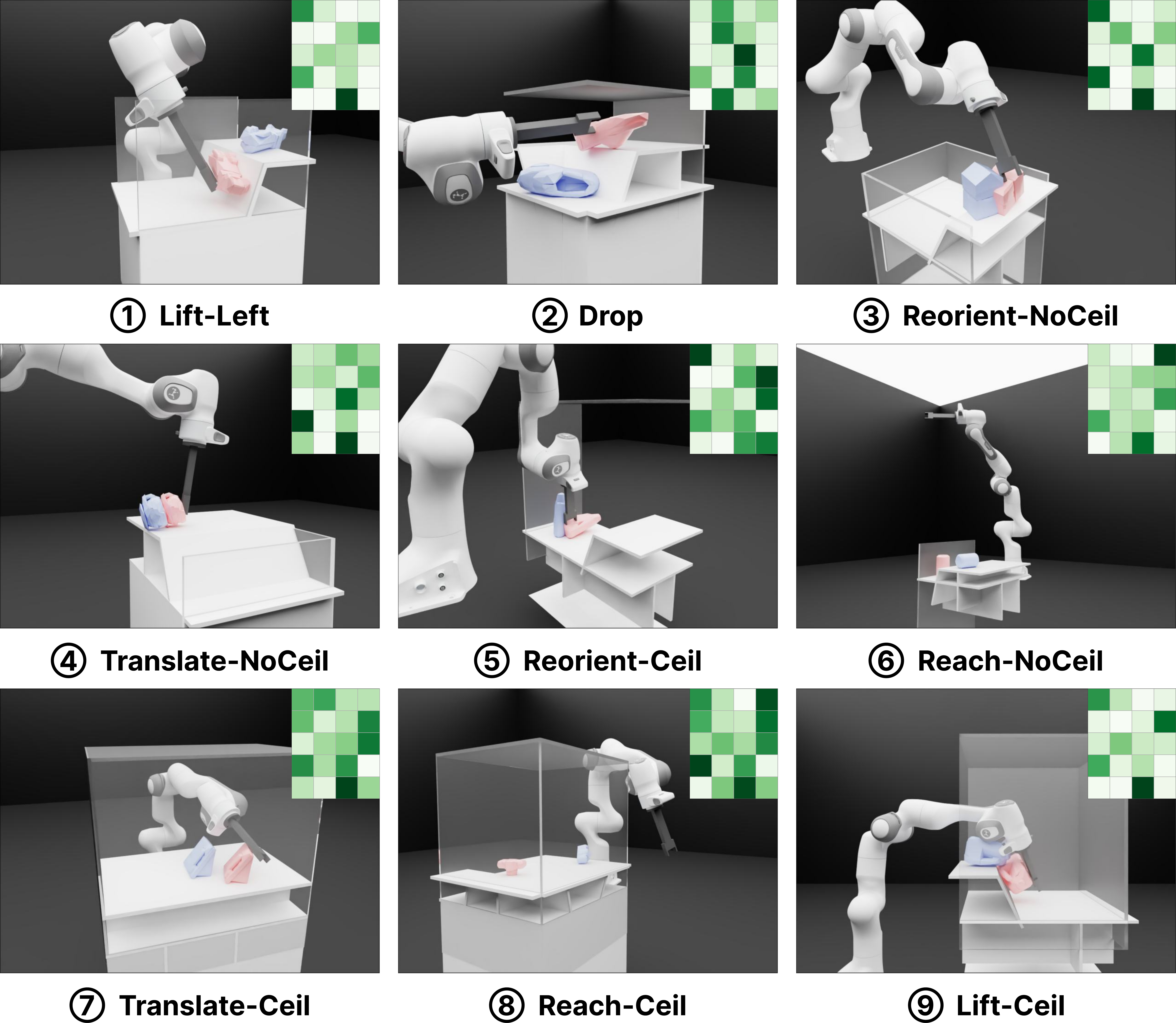}
    \caption{Illustration of how distinct scenes map to distinct module activations that yield distinct behaviors, for a 5-layer network with 4 modules. The top-right colormap shows the activation of a particular module (column) for a particular layer (row): e.g., column 2, row 1 shows the activation of module 2 for layer 1. Opacity indicates the strength of module activation. The red object denotes the current object pose, and the blue object indicates the goal object pose. Ceil and NoCeil indicate the existence of a ceiling.}
    \label{fig:domain-skill-route}
    \vspace{-2ex}
\end{figure}

\begin{figure}[t]
    \centering
    \includegraphics[width=0.7\linewidth]{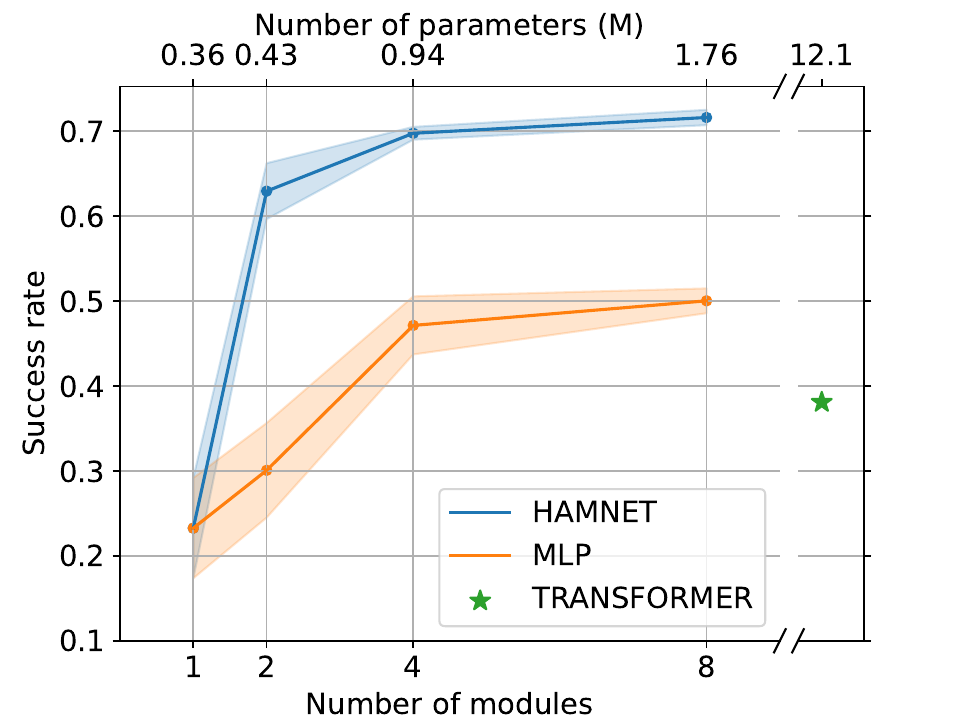}
    \caption{Success rates per architecture by parameter counts for a monolithic architecture (MLP and transformer) (x-axis, top) and number of modules for \oura{} (x-axis, bottom)}
    \label{fig:abl-scale}
    \vspace{-3ex}
\end{figure}
We refer to our modular policy architecture as \underline{H}ierarchical \underline{a}nd \underline{M}odular \underline{Net}work (\oura{}). We discovered that, when we train the policy with \oura{}, it autonomously discovers modules and their activations that correspond to distinct manipulation strategies, such as reaching, lifting, or reorienting objects (Figure~\ref{fig:domain-skill-route}). Furthermore, as shown in Figure~\ref{fig:abl-scale}, we found that \oura{} scales much more gracefully with the number of parameters compared to a standard neural network in our simulated domains.

Another challenge in generalizing across diverse environments and objects is acquiring geometric representations from high-dimensional point cloud observations. One option is to use end-to-end training, yet jointly learning to encode point clouds incurs significant memory footprint, limiting its use with large-scale GPU-based simulators~\cite{isaacgym}, which have become an essential tool for training RL policies for robotics. Pre-training representation models can mitigate this issue, but their effectiveness depends on the pretext task capturing features relevant to manipulation. Consequently, large off-the-shelf models are often unsuitable, not only from their high computational cost but also due to spurious geometric features that are irrelevant for contact-rich manipulation~\cite{zhang2021learning,cho2024corn}.

Instead, we build on CORN~\cite{cho2024corn}, which learns object representations tailored for manipulation based on a pretext task that predicts the presence and location of contact between a gripper and object. The key insight is for contact-rich manipulation, it is important to capture what forces and torques can be applied to an object in the given state, which in turn depends on the presence and location of the contact between the object and the gripper. While CORN restricts itself to gripper-object contacts, we also need object-environment contacts in our problems as we generalize over environments. So, we introduce \underline{Uni}versal \underline{CORN} (\ourr{}) that generalizes to contact affordances between two arbitrary geometries, based on a Siamese pre-training pipeline where a single encoder learns the representations for both an object and environment, and a decoder predicts the contacts between each environment point cloud patch and the object.

Our last contribution concerns domain design. To obtain a policy that generalizes to diverse environments, training environments must encompass a range of geometric constraints, while affording fast simulation for practical training. This is a non-trivial problem: unlike objects, environment assets for simulated RL training are not widely available, and manually designing them is costly. Therefore, we propose a procedural generation algorithm for constructing environments based on cuboid primitives. Online rearrangement of cuboids at different poses and dimensions yields wide coverage of geometric features that exist in the real world, such as walls, ceilings, slopes, and bumps. Further, the convex geometry of cuboids affords efficient dynamics simulation. Figure~\ref{fig:domain-skill-route} shows example environments from our environment generation algorithm.

We show that by leveraging our framework, we can train a non-prehensile manipulation policy that can operate in diverse and novel environments and objects in a data- and time-efficient manner. We train our policy entirely in simulation and zero-shot transfer to unseen environments and objects in the real world. Furthermore, we provide a simulation-based benchmark comprising 9 digital twins of real-world environments and 353 objects to serve as a benchmark for non-prehensile manipulation for general environments and objects.

\section{Related Work\label{sec:rel-work}}
\subsection{Nonprehensile Manipulation\label{sec:rel-work-np}}

\subsubsection{Planning-based approaches}

Prior works on planning-based non-prehensile manipulation use gradient-based optimization~\cite{mordatch2012contact,posa2014direct,moura2022non},
graph-based search~\cite{maeda2005astar,maeda2001dijkstra,miyazawa2005rrt,cheng2023mcts,liang2022learning,cheng2022contact}, or a hybrid of both~\cite{chen2021trajectotree,pang2023traj}.
To address the discontinuous dynamics arising from contact mode transitions,
optimization-based works employ soft contact variables~\cite{mordatch2012contact} or complementarity constraints~\cite{posa2014direct,moura2022non}.
However, due to the imprecision of smoothed contact and the difficulty of precise constraint satisfaction,
the resulting motions are difficult to realize in the real world.

On the other hand, graph-based methods~\cite{cheng2022contact,Hou_2019} can handle discrete dynamics transitions by
representing the problem with a graph, where nodes represent robot states and contact modes,
and edges encode the motion during the transition.
This enables these methods to output more physically realistic motions for real-world deployment~\cite{cheng2023mcts,liang2022learning}.
However, to make the search tractable, these works restrict the diversity of motions,
assuming quasi-static motions~\cite{cheng2022contact,Hou_2019} or predefined primitives~\citep{zito2012two,liang2022learning},
limiting them to tasks with simple motions and sparse contact-mode transitions.

Other works combine optimization and sampling to accelerate planning~\cite{chen2021trajectotree,pang2023traj},
yet remain too slow for online use due to the cost of searching large hybrid spaces with discontinuous dynamics.
Further, most planning methods require knowledge of system parameters like mass and friction,
which are difficult to estimate in real-world scenarios with varying objects and environments,
harming practical real-world deployment.

\subsubsection{Learning-based approaches}

Recent works leverage reinforcement learning (RL) to bypass the limitations of traditional planners by learning a policy that maps actions directly from sensory inputs~\cite{lowrey2018reinforcement,yuan2018rearrangement,peng2018sim,yuan2019end,ferrandis2023nonprehensile,zhou2023gug,kim2023crm,zhou2023hacman,cho2024corn,wu2024retarget}.
While this circumvents the computational cost of planning or the requirement of full system parameters,
most of these works suffer from limited generalization across object geometries,
since the policy is only trained on a single object~\cite{lowrey2018reinforcement,yuan2018rearrangement,peng2018sim,yuan2019end,ferrandis2023nonprehensile,zhou2023gug,kim2023crm}.  Recent works incorporate point-cloud inputs~\cite{zhou2023hacman,cho2024corn} or employ contact retargeting~\cite{wu2024retarget} to facilitate generalization across diverse object shapes, but none of these approaches adequately addresses the problem of generalizing across diverse environments using the full action space of the robot, as summarized in Table~\ref{tab:np-comparision}. 

\begin{table}[tb]
\caption{Comparison of generalization capabilities for non-prehensile manipulation.}
\label{tab:np-comparision}
\resizebox{\columnwidth}{!}{%
\begin{tabular}{lccc}
\hline
 & Object generalization & General action space & Environment generalization \\ \hline
HACMAN~\cite{zhou2023hacman}                          & O                     & X                     & X                          \\
CORN~\cite{cho2024corn}                            & O                     & O                      & X                          \\
Wu et al.~\cite{wu2024retarget}                       & $\triangle$           & $\triangle$                       & $\triangle$                \\
Ours                           & O                     & O                      & O                          \\ \hline
\end{tabular}%
}
\vspace{-3ex}
\end{table}

To generalize across diverse objects, HACMan~\cite{zhou2023hacman} predicts an object-centric affordance map on its point cloud. While sample-efficient, this restricts robot motion to a hand-designed poking primitive, limiting their applicability in diverse environments. CORN~\cite{cho2024corn} learns a policy over the full
joint space of the robot,
and generalizes over objects with a contact-based object representation tailored for manipulation. However, this work remains limited to a fixed tabletop due to the lack of environment representation. Wu et al.~\cite{wu2024retarget} retargets contacts from human demonstrations to determine robot actions for novel scenes, but their approach is limited to scenes similar to the original demonstration, as the actions are restricted to predefined skill sequences based on a primitive library. Like CORN, we train an RL policy over the full robot joint space to manipulate objects of general geometry. However, our approach generalizes across environments by leveraging our extended contact-based representation, \ourr{}, and a modular network architecture, \oura{}.

\subsection{Multi-task Neural Architectures \label{sec:rel-work-arch}}
In multi-task learning, a single model learns to do a family of related tasks by leveraging task synergies for improved performance and training efficiency~\cite{caruana1997mtl}. As a single model must distinguish multiple tasks, it additionally takes context variables (e.g., task IDs) as conditioning inputs. The simplest approach for multi-task learning uses a \emph{monolithic} architecture, which incorporates context inputs simply by concatenating them with network inputs. However, this design suffers from interference among tasks, as the neurons must handle multiple functions, leading to performance degradation~\cite{yu2020gradient,Jayakumar2020MultiplicativeIA}.

\begin{figure}[tb]
    \centering
    \includegraphics[width=0.38\textwidth]{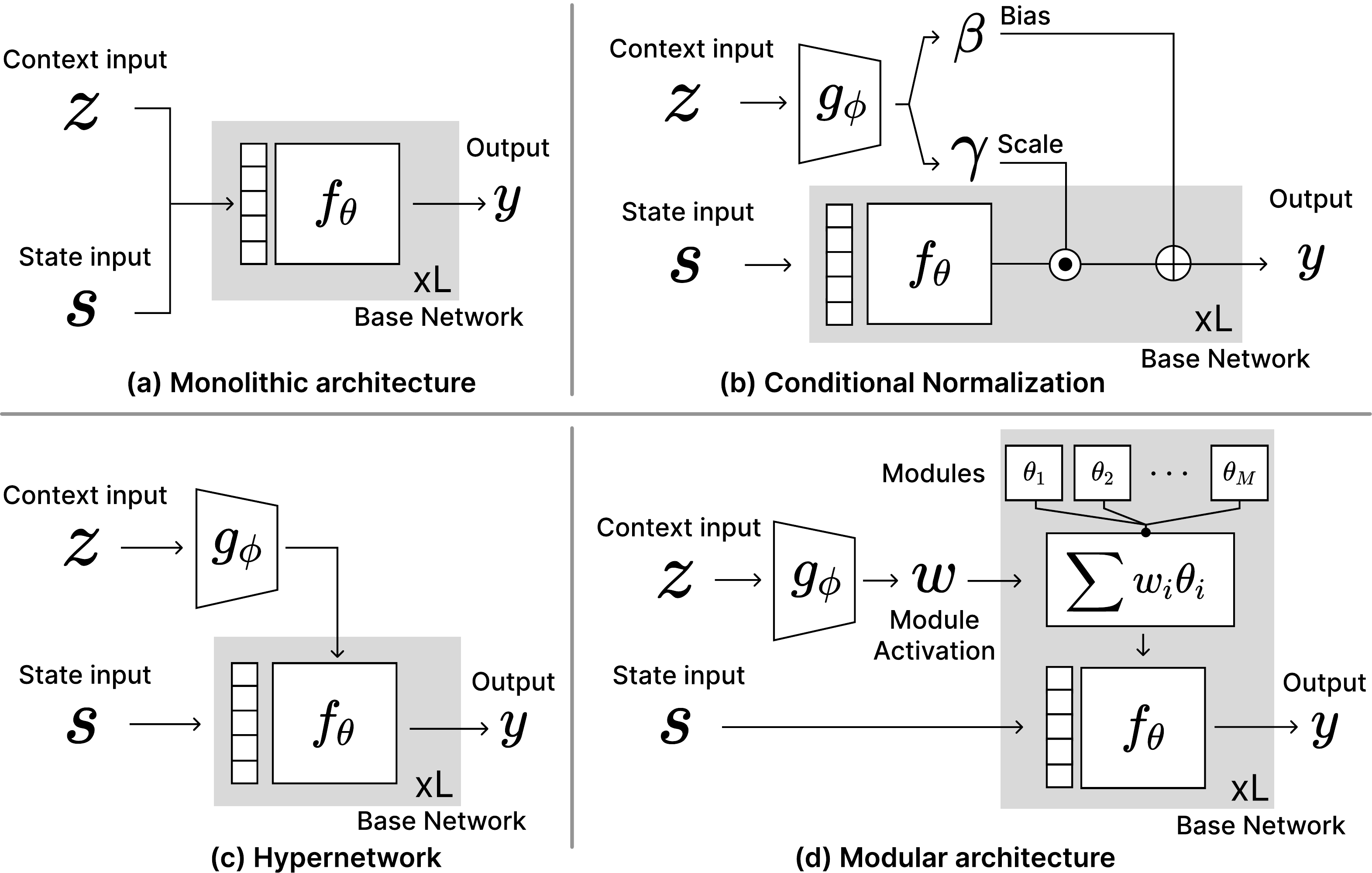}
    \caption{Conceptual illustration of how different architectures incorporate context inputs. $\bigoplus$ and $\bigodot$ denote elementwise sum and multiplication, and L denotes number of layers. The base network that computes the output is indicated by $f_\theta$, while $g_\phi$ is a separate network that determines $\theta$ from the context. In our work, we adopt a variant of the modular architecture (d).}
    \label{fig:related-work-network-diff}
    \vspace{-3ex}
\end{figure}

Recently, \emph{context-adaptive} architectures have been proposed, where a separate network $g_\phi$ determines the parameters of base network $f_\theta$ from context inputs. See, for example, Figure~\ref{fig:related-work-network-diff} (a) and (c) for a comparison of monolithic and context-adaptive architectures. This separation allows the neural network to define a different function for each context, which mitigates interference~\cite{Jayakumar2020MultiplicativeIA}.  Our architecture, \oura{}, also falls into this category.

Representative context-adaptive architectures include conditional normalization~\cite{perez2018film,peebles2023dit,mescheder2019occupancy,brohan2023rt1}, hypernetworks~\cite{ha2017hypernetworks,Jayakumar2020MultiplicativeIA}, and modular architectures~\cite{galanti2020modular,yang2020softmodule,sun2022paco}. In conditional normalization~\cite{perez2018film} (Figure~\ref{fig:related-work-network-diff}b), $g_\phi$ takes the context variable $z$ as input and applies feature-wise scale and bias $\{\gamma, \beta\} = g_\phi(z)$ to the intermediate features of the base network as $y = \gamma \odot f_\theta(x) + \beta$ where $\odot$ denotes element-wise multiplication. While effective, the expressivity of these architectures is limited to affine transforms, restricting its capacity~\cite{rebain2023attention}. In hypernetworks~\cite{ha2017hypernetworks} (Figure~\ref{fig:related-work-network-diff}c), $g_\phi$ affords broader expressivity, as it generates the entire set of base network parameters, $\theta = g_\phi(z)$, but suffer from poor training stability due to its large decision space over densely interacting parameters~\cite{ortiz2023magnitude}. 

\oura{} is an instance of modular architectures (Figure~\ref{fig:related-work-network-diff}d),
where $g_\phi$ only predicts sparse activation weights $w$ of modules to determine $\theta$. Specifically, in modular architectures, $M$ denotes the number of modules, each of which is a network parameter, and the parameters of the base network are effectively formed as a weighted combination of these modules, such as $\theta = \sum_{i=1}^{M}w_i \theta_i$.

However, prior works on modular architectures for control assume that $z$ is given by the user and is discrete, such as pre-defined task IDs~\cite{ponti2022combining,sun2022paco,yang2020softmodule,hendawy2024moore}. In contrast, to generalize to novel, real-world environments in our problem, we need to infer $z$ from sensory observations such as environment and object point clouds. To overcome this, we design geometric encoders that map high-dimensional sensory observations to $z$. 

Our architecture is most similar to Soft Modularization (SM)~\cite{yang2020softmodule}, with three key differences.
Unlike SM, which predicts \emph{connectivities} between all pairs of modules between neighboring layers, we directly predict \emph{module activations}, which simplifies computation and reduces the output dimensions from $M^2$ connections to $M$ module activations.
We additionally improve the computational efficiency by predicting all module activation weights in parallel,
instead of predicting in series while conditioning on the preceding layer's module activations.
Lastly, we incorporate a gating mechanism~\cite{ha2017hypernetworks} to enhance expressivity and boost policy performance.

\subsection{Representation learning on point clouds \label{sec:rel-work-repr}}
To accelerate RL training with high-dimensional sensory inputs, prior works use representational pre-training~\cite{lin2020learningts, farebrother2023protovalue} to bootstrap RL agents. Different pretext tasks have been proposed for this purpose, such as point completion~\cite{wang2021occo}, orientation and category estimation~\cite{chen2022visualdexterity,huang2021geodex}, or contrastive learning~\cite{xie2020pointcontrast}. Inspired by the advances in natural language processing~\cite{devlin2019bert,radford2018gpt} and image analysis~\cite{he2022mae}, recent works adopt self-supervised learning (SSL) on patch-based transformers~\cite{yu2022pointbert,chen2023pointgpt,pang2022pointmae,zhang2022pointm2ae,abouzeid2023point2vec}
for point cloud representation learning. These works reconstruct unseen geometric patches via either autoregressive prediction~\cite{chen2023pointgpt} or masking~\cite{yu2022pointbert,pang2022pointmae,zhang2022pointm2ae,abouzeid2023point2vec} to learn rich geometric representations, achieving state-of-the-art results in shape classification and segmentation~\cite{chen2023pointgpt}.

Despite their success in general-purpose vision tasks, these representations are unsuitable for robot manipulation for two reasons. First, these models attempt to predict the missing patches in a point cloud, which forces the encoder to focus on encoding information about the object's shape. However, knowing the exact shape of the object is often sufficient but unnecessary for manipulation. For example, manipulating a toy crab in Figure~\ref{fig:motivation}e does not require knowledge of the exact shape between its legs, as that area is tightly confined and cannot be contacted by the robot or the environment. Second, capturing such spurious details requires a large model, which degrades training efficiency~\cite{fang2021seed,shi2022efficacy} and policy performance~\cite{zhang2021learning}. In contrast, we extend CORN~\cite{cho2024corn} to pre-train a representation to encode contact affordances among arbitrary geometry pairs, shown to be effective for robot manipulation.

\subsection{Modularity in biological networks\label{sec:rel-work-bio}}

Modularity in biological neural networks is a key principle underlying adaptation and learning~\cite{sporns2016brain}.
In vertebrate motor systems,
\emph{muscle synergies}~\cite{Tresch1999spinal,dAvella2003synergy,Overduin2012monkey,McCrea2008organization,Dominici2011baby}
serve as modules of movement that abstract muscle control,
representing a coordinated contraction of a set of muscles to produce a desired behavior,
such as the synchronized activation of the quadriceps and hamstrings for walking~\cite{Dominici2011baby}.

Modularizing motor control in this way provides several benefits. When adapting to a particular context, the central nervous system (CNS) can dictate behaviors using sparse, low-dimensional signals that activate specific muscle groups~\cite{bizzi2013synergy}. Compared to controlling individual motor neurons, this affords rapid switching between distinct motor skills like reaching and grasping depending on the context~\cite{Overduin2015RepresentationOM,ting2007neuromechanics}. Further, synergies can be reused across behaviors, producing diverse movements such as pinch- or power-grasps from a limited set of modules~\cite{Prevete2018grasp}, which facilitates learning by recombining and adapting existing modules~\cite{clune2013modular,ellefsen2015forget}. In our work, we incorporate these principles to design our architecture.

\subsection{Skill discovery in RL\label{sec:rel-work-skill}}

In skill discovery, Unsupervised RL (URL) aims to find reusable skills using task-agnostic objectives like state coverage~\cite{curiosity,void,lexa,Skew-Fit,prototypical,rnd} or skill diversity~\cite{vic,diayn,dads,cic,lsd}. However, without task-specific priors, such intrinsically motivated methods often fail to make meaningful interactions in high-dimensional domains without engineered bias~\cite{baker2019emergent,tirinzoni2025zeroshot}. While METRA~\cite{metra} was shown to scale to pixel-based tasks, it remains prone to degenerate behaviors, such as lying still in varied poses in humanoid control domains~\cite{tirinzoni2025zeroshot}. On the other hand, our framework learns task-specific skills grounded in the training domain, yielding interpretable skills that transfer to real-world robot manipulation.

\subsection{Mitigating Spectral Bias in RL\label{sec:spectral-bias-rl}}
A central challenge in multi-task learning is enabling networks to adaptively \emph{switch} functions under subtle shifts in its input. While our work focuses on the design of \emph{network architecture}, one complementary line of work investigates improving \emph{input representations} with high-frequency encodings, such as Learned Fourier Features (LFFs)\cite{tancik2020fourier}, to enhance the network's sensitivity to sharp transitions. While promising, the effectiveness of LFFs remains inconclusive in RL, with conflicting claims on the role of different frequency components~\cite{li2021functional,yang2022overcoming,mavor2024frequency}. Moreover, LFFs lead to an exponential increase in input dimensionality~\cite{brellmann2023fourier} and may degrade training stability from the increased variance, impairing generalization and robustness to noise~\cite{mavor2024frequency}. These limitations suggest the need for an \emph{architecture-level} solution, and not just the input representation. Still, one potentially promising future direction is to see whether high-frequency embeddings can complement the modular architecture by increasing network sensitivity to distinct contexts when determining module activations.

\begin{figure}[t]
    \centering
    \includegraphics[width=0.8\linewidth]{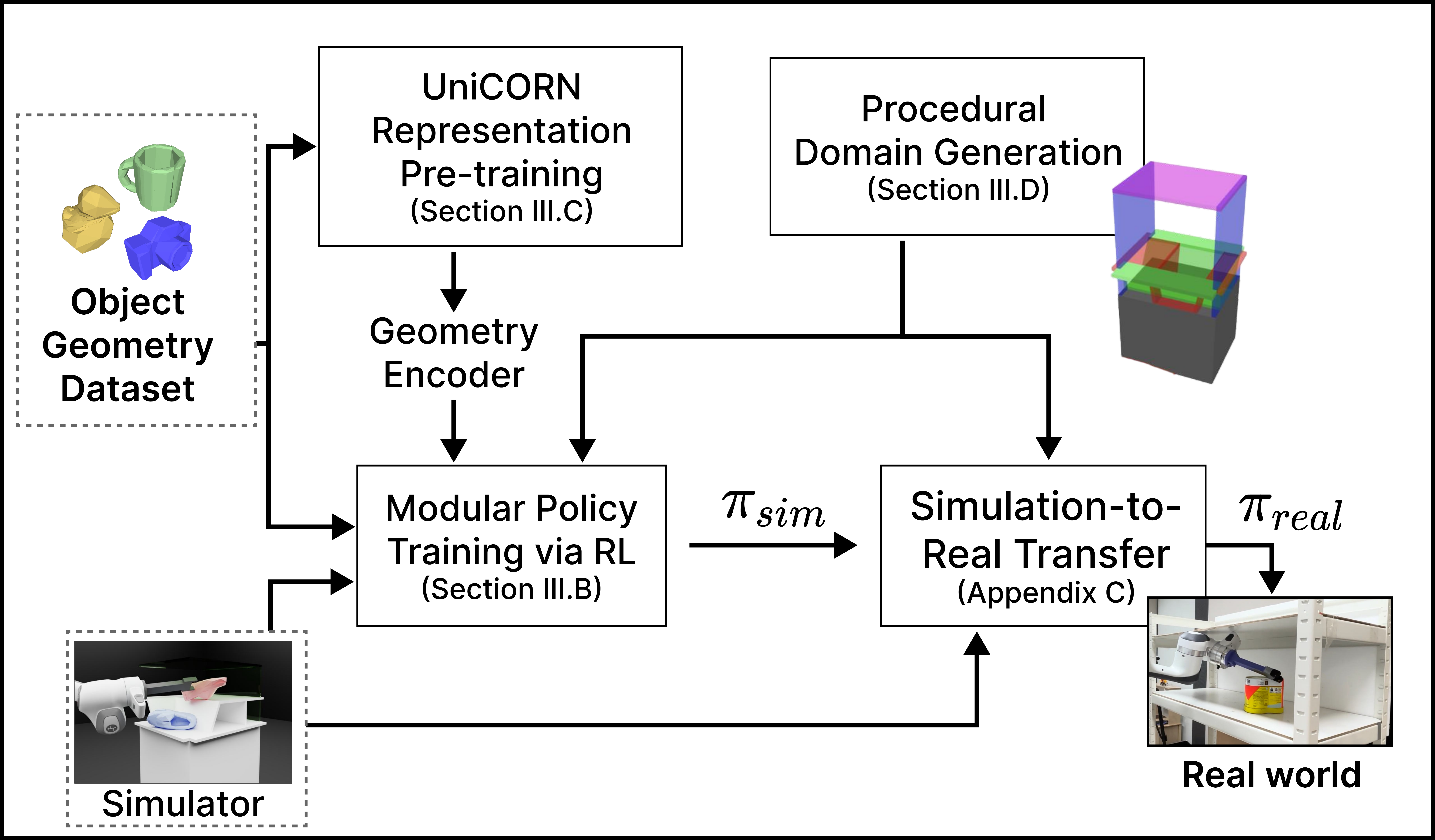}
    \caption{Overall method overview. Our framework consists of four main components: a modular policy trained with RL, contact-based representation pre-training, a procedural domain generation scheme for environment geometries, and a simulation-to-real transfer method for real-world deployment. Dashed blocks indicate external inputs.}
    \label{fig:overall-method}
    \vspace{-4ex}
\end{figure}

\begin{figure*}[th]
    \centering
    \includegraphics[width=0.9\textwidth]{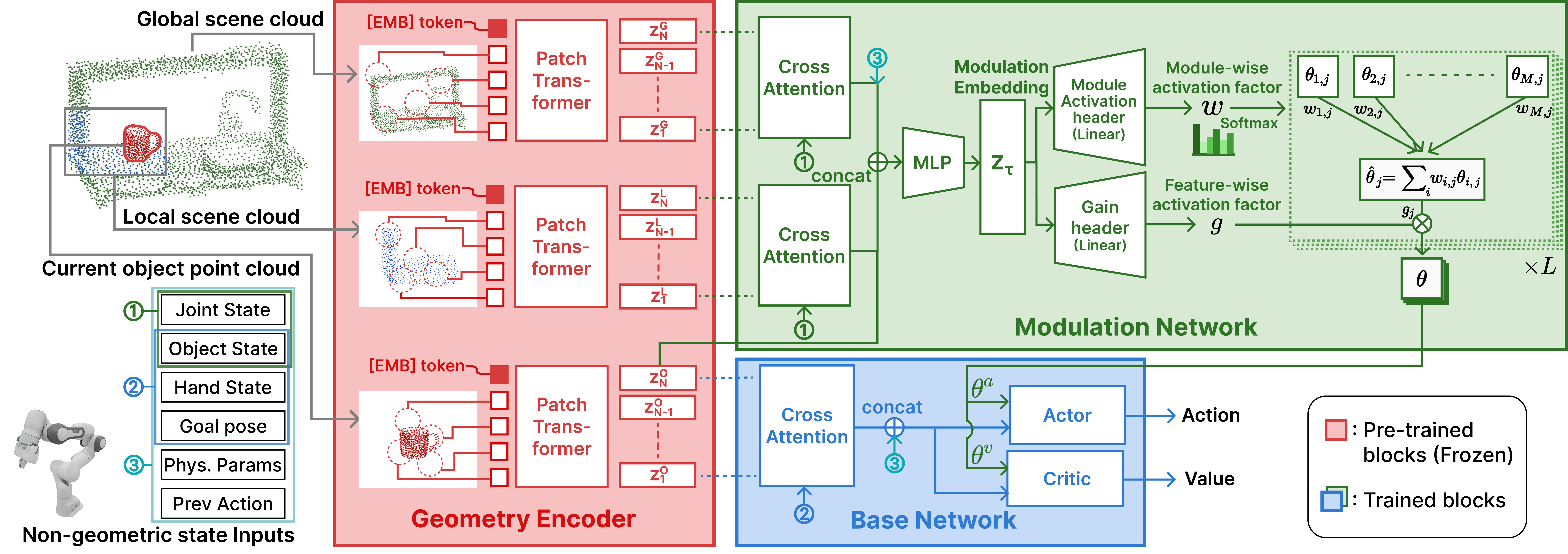}
    \caption{Overall architecture. Our model comprises three components -- the geometry encoder (red), the modulation network (green), and the base network (blue). The geometry encoder embeds the point clouds, and the modulation network maps the embeddings and non-geometric state inputs to the base network's parameters $\theta$. Conditioned on $\theta$, the base network maps the state inputs and object geometry to actions and values. Input groups tagged with different numbers ({\large \textcircled{\small 1}},  {\large \textcircled{\small 2}} and {\large \textcircled{\small 3}}) indicate sets of non-geometric state inputs fed into different network parts. The inputs to cross-attention layers are concatenated and tokenized by a two-layer multi-layer perceptron (MLP).}
    \label{fig:overall-arch}
    \vspace{-2ex}
\end{figure*}

\section{Methodology\label{sec:method}}

We consider the non-prehensile manipulation problem, where a robot arm with a fixed base moves an object to a target pose in environments of general geometry, e.g., kitchen sinks, cabinets, and drawers~(Figure~\ref{fig:motivation}). We model this problem as a Markov Decision Process (MDP), represented as a tuple $(S, A, P, r, \gamma)$ denoting state space $S$, action space $A$, state transition model $P(s_{t+1}|a_{t},s_{t})$, reward model $r(s_t, a_t, s_{t+1})$, and discount factor $\gamma$. Our objective is to obtain policy $\pi$ that maximizes the return $R_t=\mathbb{E}_{a_t\sim\pi(\cdot|s_t)}[\sum\gamma^{t}r(s_t,a_t,s_{t+1})]$ via a sequence of non-prehensile actions.

Figure~\ref{fig:overall-method} presents an overview of our framework.
We leverage deep RL in a parallel GPU-based simulation~\cite{isaacgym} to train a modular policy (Section~\ref{sec:method-modular-arch})
using the pre-trained point cloud representation (Section~\ref{sec:method-representation}) on procedurally generated domains (Section~\ref{sec:proc-gen}).
We distill the resulting policy for real-world deployment via teacher-student distillation (Appendix~\ref{sec:sim2real-transfer}).
All pre-training, policy training, and distillation stages happen entirely in a simulation. 

\subsection{MDP Design\label{sec:mdp-design}} 

Our state space $S$ consists of robot joint state $x^q_t$, end-effector pose $x_t^{EE}$, object geometry $G_o$, environment geometry $G_e$, and goal pose $T_g$. In the simulation, the agent additionally receives physics parameters $\nu$ and object state $x^o_t$. We represent all poses as 3D translation and 6D orientation to facilitate learning~\cite{zhou2019continuity}, and the goal is given as a relative pose from the current object pose. Object and scene geometries are given as surface-sampled point clouds.

Our action space $A$ consists of joint residuals $\Delta q\in\mathbb{R}^7$
and controller gains, parameterized by proportional gains $k_p \in\mathbb{R}^7$
and the damping ratio $\rho \in \mathbb{R}^{7}$ that maps to the damping gain $k_d$ as $\rho \sqrt{k_p}$, following~\cite{vice2019martin,kim2023crm}.
The resulting torque for each joint is computed as $\tau=k_p\Delta q - k_d\dot{q}$. While prior works adopt Cartesian-space actions~\cite{zhou2023hacman,cho2024corn}, we adopt joint-space actions, which enables direct control of individual robot links to avoid collisions against the environment during manipulation.

The reward $r(s_t, a_t, s_{t+1})$ in our domain is defined as a sum of the task success reward $r_{s}$, goal-reaching reward $r_{r}$ and the contact-inducing reward $r_{c}$: $r=r_{s}+\lambda_{r}r_{r}+\lambda_{c}r_{c}$,
where $\lambda_{r}$ and $\lambda_{c}$ are scaling coefficients for the respective rewards. Since $r_{s}=\mathds{1}_{suc}$ is sparsely given, we incorporate shaping rewards $r_{r}$ and $r_{c}$ as potential functions of the form $\gamma\phi(s') - \phi(s)$ with the discount factor $\gamma \in [0,1)$, which preserves policy optimality~\cite{ng1999potential}. Specifically, we have $\phi_{r}(s) = -\log{(c_{g} \cdot d_{o,g}(s)+1)}$ for $r_{r}$, and $\phi_{c}(s) = -\log{(c_{r} \cdot d_{h,o}(s)+1)}$ for $r_{c}$, where $c_g,c_r \in \mathbb{R}$ are scaling coefficients for the distance-based potential functions; $d_{o,g}(s)$ is the relative distance between the current object and the goal pose, based on the bounding-box distance~\cite{allshire2022transferring}; $d_{h,o}(s)$ is the hand-object distance between the object and the tip of the end-effector. Task success is achieved when the object's pose is within 0.1m and 0.1 radians of the target pose. The episode terminates if (1) the object reaches the goal, (2) the object is dropped from the workspace, or (3) the episode reaches the timeout of 300 simulation steps. Table~\ref{tab:mdp_summary} summarizes our MDP design, and details on reward coefficients are in Table~\ref{tab:reward-params}.

\newcommand{\zg}{z^{(G)}}
\newcommand{\zl}{z^{(L)}}
\newcommand{\zo}{z^{(O)}}

\subsection{\textnormal{\oura}-based architecture \label{sec:method-modular-arch}}
Our architecture, shown in Figure~\ref{fig:overall-arch},
consists of three main components:  the geometry encoder (red), modulation network (green), and base network (blue). Our proposed modular architecture, \oura, consists of the modulation and base networks. Since we use PPO, an actor-critic algorithm, our base network outputs both value and action. 

The geometry encoder processes three types of point cloud inputs: the \emph{global} scene cloud, capturing the overall geometry of the scene; the \emph{local} scene cloud, detailing the nearby scene that surrounds the object; and the \emph{object} point cloud, representing its surface geometry. Each cloud is patchfied, tokenized, and embedded by the pre-trained geometry encoder (Section~\ref{sec:method-representation}), yielding latent geometric embeddings $\zg_{\text{1:N}}$, $\zl_{\text{1:N}}$, $\zo_{\text{1:N}}$,
for global, local, and object embeddings, respectively. Details on point cloud acquisition are in Appendix~\ref{sec:cloud-sample}.

The role of the modulation network (Figure~\ref{fig:overall-arch}, green) is to output the parameters of the base network, $\theta$. It takes the geometry embeddings $\zg_{1:N},\zl_{1:N},\zo_N$ and non-geometric states ({\large \textcircled{\small 1}} and {\large \textcircled{\small 3}}) as input. To extract scene geometry information relevant to the policy's current state, we apply cross-attention on the scene geometry embeddings $\zg_{1:N}$ and $\zl_{1:N}$, using the current robot and object states {\large \textcircled{\small 1}} as queries. The resulting vector is concatenated with object geometry embedding $z_N^{O}$ and full non-geometric state inputs {\large \textcircled{\small 3}}, and passed through an MLP to predict $z_\tau$, the modulation embedding. Finally, the module activation and gain headers map $z_\tau$ to module activation weights $w$ and gating values $g$ respectively, for the $L$ base network layers. 

We then use $w$ and $g$ to build the base network parameters $\theta$. For each layer $j \in [1\dots L]$, $w=\{w_{i,j}\}_{j=1}^{L} \in \mathbb{R}^{L \times M}$ act as M module-wise weighting coefficients, passed through softmax to ensure $\sum_{i=1}^{M} w_{i,j}=1$. The gating factor $g=\{{g_j}\}_{j=1}^{L} \in \mathbb{R}^{L \times D_j}$ is a feature-wise multiplier for each layer, with $D_j$ denoting the number of output dimensions of layer $j$. Together, $\theta$ is constructed as a weighted composition of modules followed by gating, such that $\theta = \{(\sum_{i=1}^{M} w_{i,j} \theta_{i,j}) \odot g_j\}_{j=1}^{L}$.

The base network (Figure~\ref{fig:overall-arch}, blue) comprises actor and critic networks, where each network is an MLP. To produce the input for the base network, we first process the object embedding $\zo_{1:N}$ via cross-attention against input group {\large \textcircled{\small 2}}, then concatenate the result with input group {\large \textcircled{\small 3}}. The actor network outputs the action, and
the critic network outputs the state values, but instead of a single scalar value, it uses three heads to predict the value for each reward component in our domain: $r_{s}$, $r_{r}$, and $r_{c}$. Since summing the rewards conflates the contributions from different reward terms, splitting the critic into multiple headers helps decrease the difficulty of value estimation~\cite{fatemi2022orchestrated,macglashan2022value}. When training the actor network, we sum the advantages across reward terms to compute the policy gradients.

Note that since the base network has both the actor and critic, we keep separate sets of modules for each of them, denoted $\{\theta_{i,j}^{(a)}\}_{i=1}^{M}$, and $\{\theta_{i,j}^{(v)}\}_{i=1}^{M}$ for layer $j$. Our module activation weights and gating factor also consist of weights for a value and action, $w=\{w^{(v)},w^{(a)}\}$ and $g=\{g^{(v)},g^{(a)}\}$. To make a prediction, the base network gets instantiated twice for actor and critic; in the former case, the network uses the weight $\theta^{(a)}_j = \sum_{i=1}^M w_{i,j}^{(a)} \theta_{i,j} \odot g^{(a)}_j\}_{j=1}^{L}$, and in the latter, the network uses $\theta^{(v)}_j = \sum_{i=1}^M w_{i,j}^{(v)} \theta_{i,j} \odot g^{(v)}_j\}_{j=1}^{L}$. These details are omitted in the figure for brevity.

\begin{figure}[t]
     \centering
        \includegraphics[width=\linewidth]{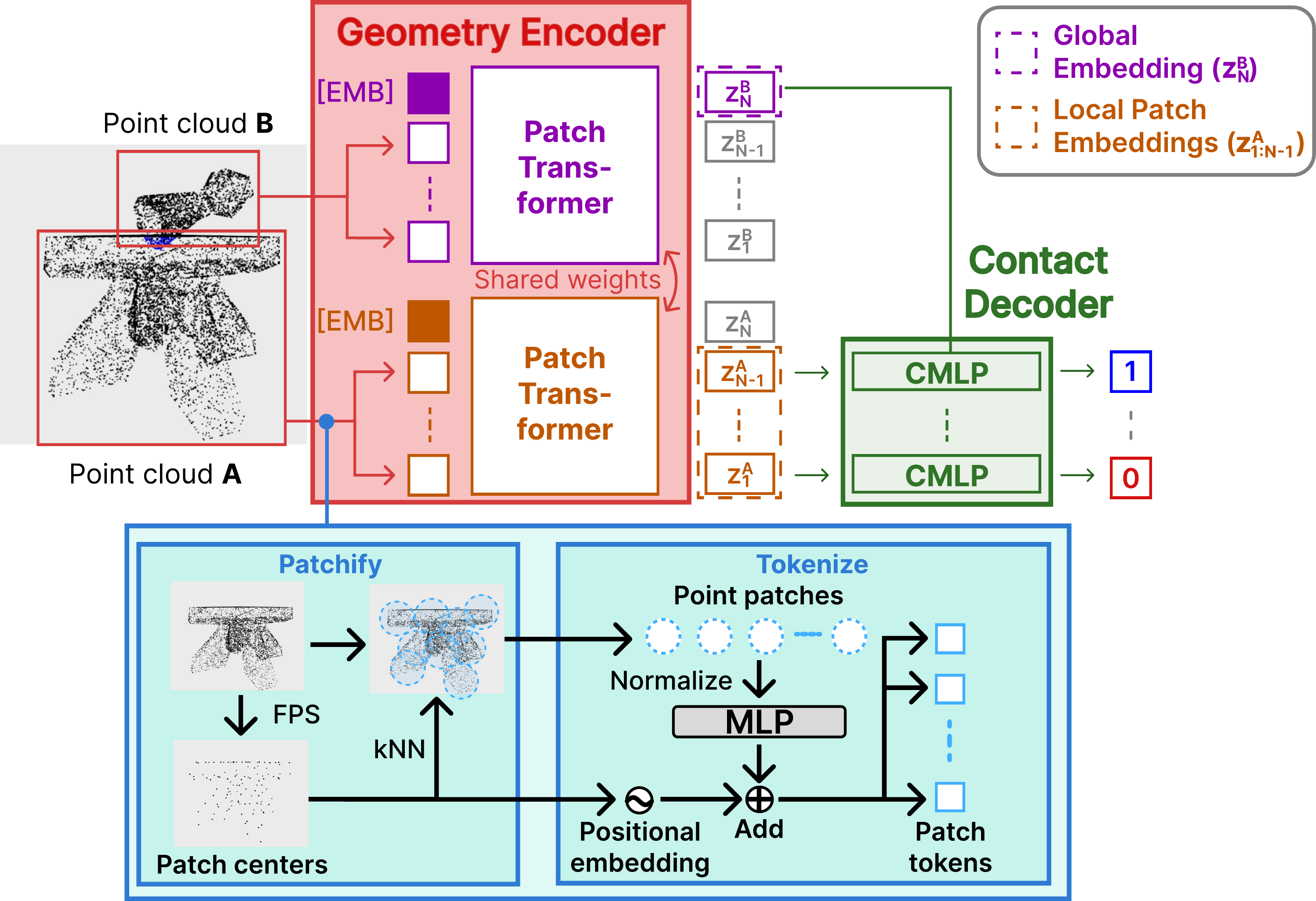}
         \caption{Our pre-training architecture consists of a geometry encoder (red) and a contact decoder (green). The same geometry encoder operates on each point cloud A and B in a Siamese fashion to produce local patch embeddings $z^A_{1:N-1}$, $z^B_{1:N-1}$ and global embeddings $z^A_{N}$, $z^B_{N}$. The contact decoder (green) predicts contact between each patch $z^A_{i} \in z^{A}_{1:N-1}$ and $z^B_{N}$. The bottom block details the procedure to patchify and tokenize point clouds.
         \label{fig:pretraining_network}
         }
     \vspace{-2ex}
\end{figure}
\subsection{
Training \textnormal{\textsc{UniCORN}}\label{sec:method-representation}
}
We design our representation pre-training task on estimating the presence
and location of contact between two point clouds, A and B.

\subsubsection{Pre-training data generation \label{sec:pre-training-data-gen}}
To acquire data for pre-training, we generate a dataset containing pairs of objects represented as point clouds, and contact labels indicating the presence and location of contact. Using the objects from DexGraspNet dataset~\cite{wang2023dexgraspnet}, we generate the data by (1) sampling \emph{near-contact} object configurations,  (2) creating the point clouds by sampling points from the surface of each object, and (3) labeling contact points based on whether they fall within the other object. To account for possible scale variations between geometries, we sample the point clouds at varying densities and scales during this process. For details on the data generation pipeline, see Appendix~\ref{sec:data-gen-unicorn}.

\subsubsection{Network Architecture \label{sec:pre-training-arch}}
Figure~\ref{fig:pretraining_network} shows our pretraining network architecture, comprising the geometry encoder and the contact decoder. The encoder takes the point clouds of objects A and B, denoted $x_{A},x_{B}$ as inputs, mapping the patch-wise tokens from $x_A$ and $x_B$ and a learnable \texttt{[EMB]} token to local patch embeddings $z_{1:N-1}^{A}$, $z_{1:N-1}^{B}$ and global embeddings $z_{N}^{A}$,$z_{N}^{B}$. Afterward, the decoder takes $(z_i^{A},z_{N}^{B})_{i=1}^{N}$ and predicts the presence of contact at each of $i$-th local patch of object A with object B. The overall network is trained via binary cross-entropy against the patch-wise contact labels. During training, we alternate the roles of A and B (i.e., A-B and B-A) to ensure that we also use the global embedding of A and predict the contact at a patch of B.

Figure~\ref{fig:pretraining_network} (bottom) shows the procedure to tokenize the point clouds.
In line with previous patch-based transformer architectures for point clouds~\cite{pang2022pointmae,chen2023pointgpt,cho2024corn}, we first patchify the point cloud by gathering neighboring points from representative center points. These center points are selected via farthest-point sampling (FPS), and the points comprising the patches are determined as the k-nearest neighbors (kNN) of the patch center. These patches are normalized by subtracting their center coordinates, and a small MLP-based tokenizer~\cite{cho2024corn} embeds the shape of each patch. Afterward, we add sinusoidal positional embeddings of the patch centers to the patch tokens to restore the global position information of each patch.

\subsection{Procedural domain and curriculum generation\label{sec:proc-gen}}
\begin{figure}[t]
\centering
\includegraphics[width=0.9\linewidth]{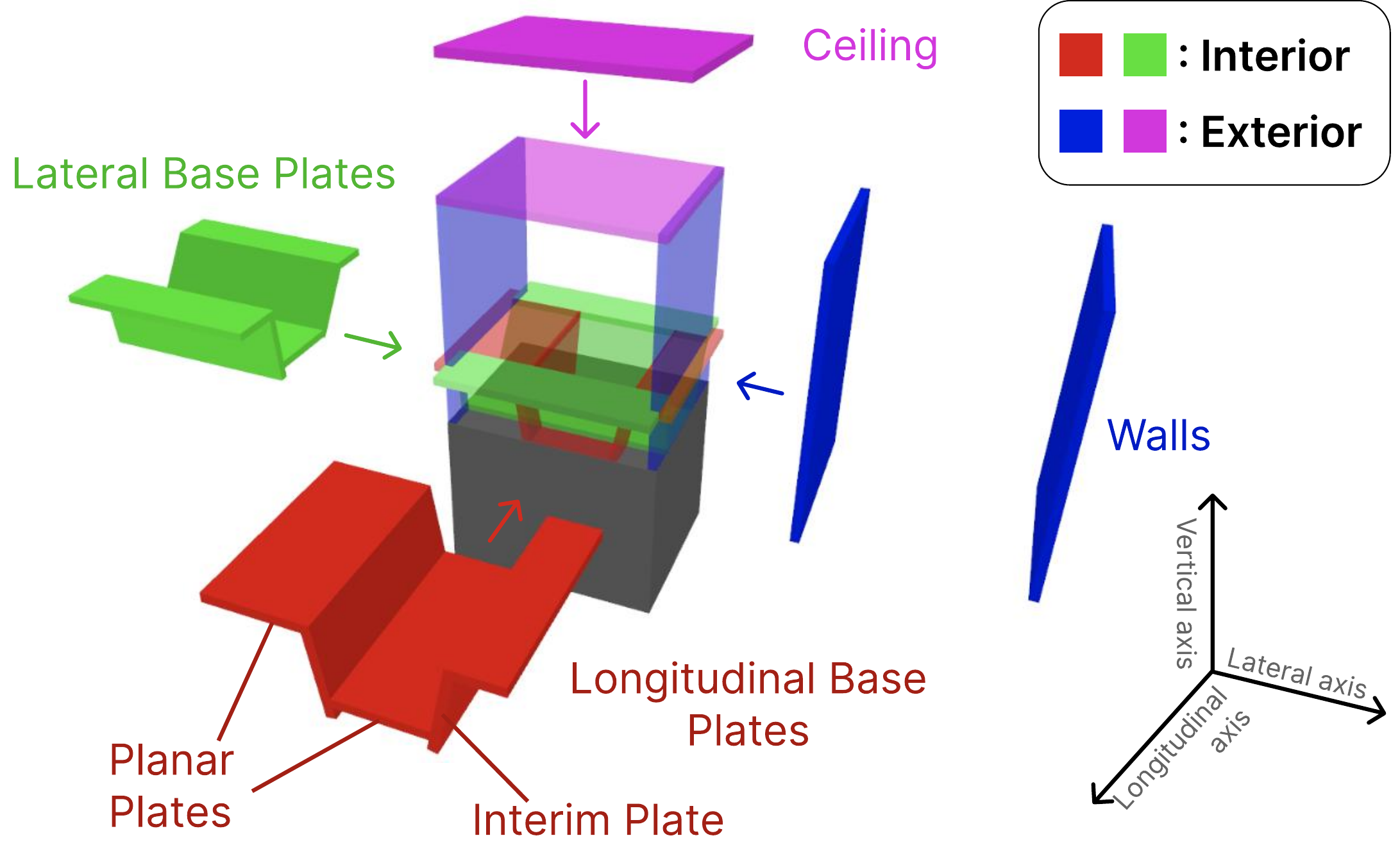}
         \caption{Our pipeline for environment generation composes different environmental factors, such as walls, ceilings, and plates at different elevations for each axis, to construct geometrically diverse environments.
         \label{fig:proc-gen-schem}
         }
         \vspace{-2ex}
\end{figure}

To create diverse environments and support curriculum learning for training our policy,
we develop a procedural generation scheme for constructing environments as a composition of cuboidal primitives.
Since we construct environments by dynamically rearranging existing geometric entities,
it integrates well with most GPU-based simulators~\cite{isaacgym,brax}
that prohibit spawning new assets after initialization.

Our procedural pipeline, shown in Figure~\ref{fig:proc-gen-schem}, comprises two main components: domain \emph{interior} and \emph{exterior} generation.
The \emph{interior} includes \emph{planar} and \emph{interim} plates arranged laterally and longitudinally,
where \emph{planar} plates form elevated surfaces and \emph{interim} plates form sloped ramps.
Their dimensions, elevations, and angles are randomly sampled to produce diverse topographies, yielding features like bumps, valleys, and steps (Figure~\ref{fig:domain-skill-route}).
The \emph{exterior} consists of walls and ceilings that impose accessibility constraints, where their presence, height, and ceiling type (\emph{nominal} or \emph{tight}) are randomly sampled.
The proportion of ceiling types controls the difficulty of workspace accessibility, since the \emph{nominal} ceiling is generated with sufficient clearance, whereas \emph{tight} ceilings leave a narrow margin relative to the object's height.
Details on the procedural generation pipeline are in Appendix~\ref{sec:domain-gen-detail}.

As our procedural generation pipeline is fully parameterized,
the sampling distributions of environmental parameters can be dynamically adjusted during training.
This enables \emph{curriculum learning}, where task complexity is incrementally increased throughout training. 
Specifically, we employ a curriculum for \emph{robot initialization} and \emph{ceiling types}.
To facilitate this, we additionally introduce two types of robot initializations: \emph{near} and \emph{random}. 
In the \emph{near} configuration, the end-effector begins within a 0.1m radius of the object based on collision-free inverse kinematics solutions from CuRobo~\cite{curobo}; 
in the \emph{random} configuration, a collision-free joint configuration is uniformly sampled within the robot's joint limits.
Early in training, we preferentially sample \emph{near} initializations and \emph{nominal} ceilings to encourage interaction with the object. 
As training progresses, we linearly increase the proportion of \emph{random} initializations and \emph{tight} ceilings, 
encouraging the policy to develop obstacle-aware maneuvers for approaching objects from arbitrary configurations.

\section{Experimental Results\label{sec:result}}
\subsection{Overview\label{sec:result-overview}}

Our goal is to evaluate the following claims:
(1) our modular architecture, \oura{}, affords data-efficient training
for a policy that generalizes over large domain diversity,
compared to monolithic or hypernetwork architectures;
(2) our contact-based representation, \ourr{},
affords data-efficient training
for a robot manipulation policy
in geometrically rich domains compared to an off-the-shelf self-supervised representation;
(3) our framework affords real-world transfer
and generalization to novel environment geometries despite
only training in a simulator with synthetic environments.

\begin{table}[t]
\centering
\caption{Comparison between baselines regarding architecture and representation.}
\resizebox{1.0\linewidth}{!}{
\begin{tabular}{l|c|c} 
\toprule
\multicolumn{1}{c|}{Model Name} & Model Architecture & \multicolumn{1}{c}{Representation}  \\ 
\hline
\textsc{UniCORN-HAMnet (ours)}                   & \oura{}     & \textsc{UniCORN}                        \\
\textsc{UniCORN-Hyper}             & Hypernetwork       & \textsc{UniCORN}                        \\
\textsc{UniCORN-SM}                & Soft-Modularization~\cite{yang2020softmodule}       & \textsc{UniCORN}                        \\
\textsc{UniCORN-Transformer}              & Transformer     & \textsc{UniCORN}                        \\
\textsc{UniCORN-Mono}              &  MLP         & \textsc{UniCORN}                        \\
\textsc{PointGPT-HAMnet}               & \oura{}     & PointGPT~\cite{chen2023pointgpt}                             \\
\textsc{E2E-HAMnet} & \oura{}     & End-to-end                        \\
\bottomrule
\end{tabular}
}
\label{tab:baseline}
\vspace{-2ex}
\end{table}

To evaluate our claims, we compare the performance of our proposed model (\ourr{}-\oura{}) with the baselines summarized in Table~\ref{tab:baseline}. These baselines explore alternative choices in network architecture or representation.
\textsc{UniCORN-Hyper} uses a hypernetwork~\cite{littwin2019hyper,sarafian2021recomposing} to predict base network parameters.
\textsc{UniCORN-SM} is a variant of a modular architecture using Soft Modularization~\cite{yang2020softmodule}.
We include two variants of standard monolithic architectures, \textsc{UniCORN-Mono} and \textsc{UniCORN-Transformer}, respectively using an MLP and a transformer.
\textsc{PointGPT}-\oura{} and \textsc{E2E}-\oura{} considers alternative choices in the representation, where the former replaces the pretrained \ourr{} with a PointGPT encoder~\cite{chen2023pointgpt}, while the latter jointly trains the representation model end-to-end.
All architectures are configured to have a similar number of trainable parameters up to the architectural constraints.
Additional details on the baselines are in Appendix~\ref{sec:baseline-arch}.

\subsection{Simulation experiment\label{sec:sim-exp}}
To train our policy, we use a Franka Research 3 (FR3) arm manipulating a subset of 323 objects from DexGraspNet dataset~\cite{wang2023dexgraspnet}
on the procedurally generated environments as in Section~\ref{sec:proc-gen}.
We train each baseline using PPO~\cite{schulman2017ppo} with identical hyper-parameters,
spanning 2 billion environment interactions across 1024 parallel environments in Isaac Gym~\cite{isaacgym}.
Detailed hyperparameters for policy training are described in Appendix~\ref{sec:baseline-arch}.
We consider two metrics: data efficiency and time efficiency.

\begin{figure}[t]
    \centering
    \includegraphics[width=0.9\linewidth]{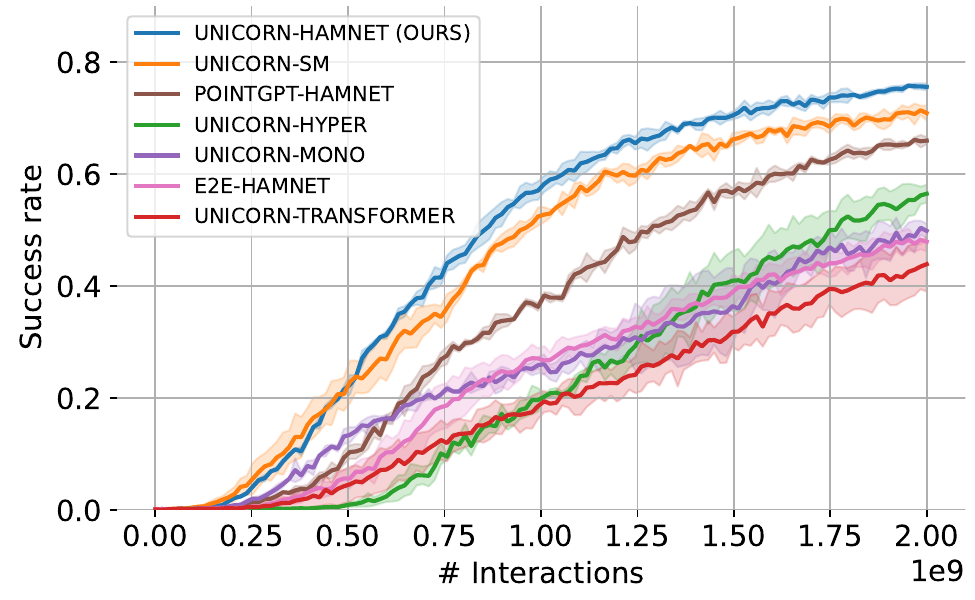}
    \caption{Training progression. For each baseline, we show the mean (solid) and standard deviation (transparent) of success rates across three seeds. The interaction steps are aggregated across 1024 parallel environments.}
    \label{fig:quantitative_training_plot}
    \vspace{-2ex}
\end{figure}

To support our claim on training efficiency, we consider the training progression plot in Figure~\ref{fig:quantitative_training_plot}.
Overall, modular architectures (\textsc{UniCORN-HAMNet} and \textsc{UniCORN-SM}) achieve the best data efficiency,
with the mean success rates of 75.6\% and 70.9\% after training.
In contrast, monolithic architectures show lower performance,
regardless of whether conditioning is given by concatenation (\textsc{UniCORN-Mono}, 49.8\%) or self-attention (\textsc{UniCORN-Transformer}, 43.9\%).
\textsc{UniCORN-Hyper} performs best among non-modular architectures at 56.5\%,
indicating the adaptivity of the network expedites policy training.
\textsc{UniCORN-SM} (70.9\%) further improves over hypernetworks, as its modularity affords
reuse of network modules and reduces the learning complexity by predicting sparse module activations rather than parameters of individual neurons.
Lastly, \textsc{UniCORN-HAMNet} (75.6\%) outperforms \textsc{UniCORN-SM} 
from the increased expressivity of the gating mechanism.
To evaluate the representational efficacy of \ourr{}, we also compare \textsc{UniCORN-HAMNet} to \textsc{E2E-HAMNet} and \textsc{PointGPT-HAMNet}. End-to-end training (\textsc{E2E-HAMNet}) degrades performance (47.9\%), emphasizing the utility of pre-training; while \textsc{PointGPT-HAMNet} performs better (66.0\%), it still underperforms \textsc{UniCORN-HAMNet}
due to the overhead from spurious geometric details and increased embedding dimensions. 

\begin{figure}[tb]
    \centering
    \includegraphics[width=0.85\linewidth]{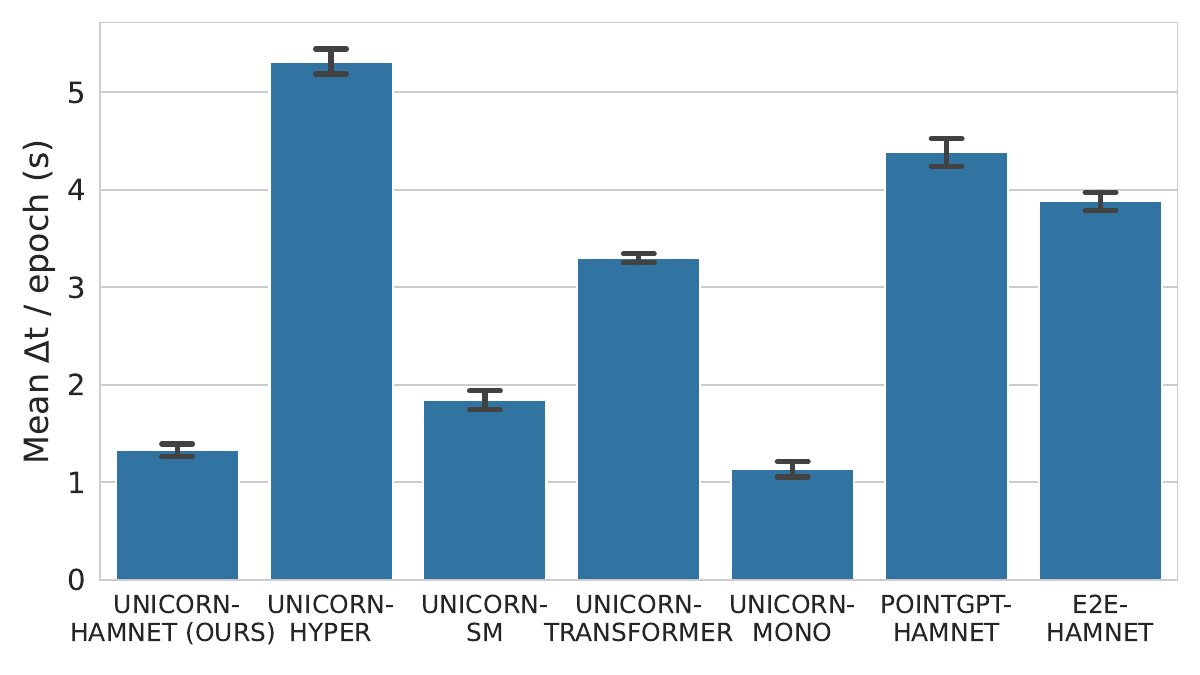}
    \caption{Per-epoch training time comparison across all baselines, measured on identical hardware (NVIDIA A6000). Error bars represent two standard deviations.}
    \label{fig:dt_bar_plot}
    \vspace{-2ex}
\end{figure}

Since all our training happens in simulation, training time is also an important factor. To evaluate the time-efficiency of each baseline, we measure the per-epoch training time in Figure~\ref{fig:dt_bar_plot}.
While the monolithic MLP architecture (\textsc{UniCORN-Mono}) is the fastest (1.12s) due to its simplicity,
modular architectures (\textsc{UniCORN-HAMNet} and \textsc{UniCORN-SM}) follow closely at just 1.33s and 1.84s,
which shows that determining module-level activation adds minimal overhead;
between the two, \textsc{UniCORN-HAMNet} achieves faster training than \textsc{UniCORN-SM} from the streamlined prediction of module activations.
In contrast, \textsc{UniCORN-Hyper} takes the longest (5.31s) due to the cost of predicting the full set of base network parameters.
To contextualize the overhead, a standard transformer (\textsc{UniCORN-Transformer}) takes around 3.30s.
We also compare with representational baselines: \textsc{PointGPT-HAMNet} and \textsc{E2E-HAMNet}.
The large encoder of \textsc{PointGPT-HAMNet} significantly lags training,
averaging 4.40s per epoch, about 3.3 times slower than \textsc{UniCORN-HAMNet}.
While \textsc{E2E-HAMNet} uses the same encoder architecture as \textsc{UniCORN-HAMNet}, it suffers from 2.9$\times$ slower training due to the overhead of co-training the representation model.

\subsection{Real world experiment\label{sec:real-world}}

\begin{table}[th    ]
\vspace{-2ex}
\centering
\caption{Results on 9 unseen real-world domains.}
\label{tab:real-result}
\resizebox{0.98\linewidth}{!}{%
\begin{tabular}{c|c|l|c|cl}
\hline
\textbf{Domain} &
  \textbf{Object} &
  \textbf{Success rate} &
  \textbf{Domain} &
  \multicolumn{1}{c|}{\textbf{Object}} &
  \textbf{Success rate} \\ \hline
\multirow{2}{*}{Cabinet} & Bulldozer    & 4/5 & \multirow{2}{*}{Top of cabinet} & \multicolumn{1}{c|}{Bulldozer} & 3/5                  \\ \cline{2-3} \cline{5-6} 
                               & Heart-Box       & 3/5 &                               & \multicolumn{1}{c|}{Crab}       & 4/5                  \\ \hline
\multirow{2}{*}{Sink}          & Bulldozer    & 5/5 & \multirow{2}{*}{Basket}   & \multicolumn{1}{c|}{Bulldozer}  & 3/5                  \\ \cline{2-3} \cline{5-6} 
                               & Angled Cup   & 4/5 &                               & \multicolumn{1}{c|}{Heart-Box}  & 5/5                  \\ \hline
\multirow{2}{*}{Drawer}        & Bulldozer    & 4/5 & \multirow{2}{*}{Grill}        & \multicolumn{1}{c|}{Bulldozer}  & 5/5                  \\ \cline{2-3} \cline{5-6} 
                               & Pencil case  & 3/5 &                               & \multicolumn{1}{c|}{Dino}       & 4/5                  \\ \hline
\multirow{2}{*}{Circular bin}  & Bulldozer    & 4/5 & \multirow{2}{*}{Flat}         & \multicolumn{1}{c|}{Bulldozer}  & 5/5                  \\ \cline{2-3} \cline{5-6} 
                               & Pineapple    & 3/5 &                               & \multicolumn{1}{c|}{Nutella}    & 3/5                  \\ \hline
\multirow{2}{*}{Suitcase} &
  Bulldozer &
  4/5 &
  \multirow{2}{*}{\begin{tabular}[c]{@{}c@{}}Total \end{tabular}} &
  \multicolumn{1}{l}{\multirow{2}{*}{}} &
  \multicolumn{1}{c}{\multirow{2}{*}{78.9\%}} \\ \cline{2-3}
                               & Candy Jar & 5/5 &                               & \multicolumn{1}{l}{}            & \multicolumn{1}{c}{} \\
\hline
\end{tabular}
}
\vspace{-1ex}
\end{table}

\begin{figure}[t]
    \centering
    \begin{subfigure}[t]{0.49\linewidth}
        \includegraphics[width=\linewidth]{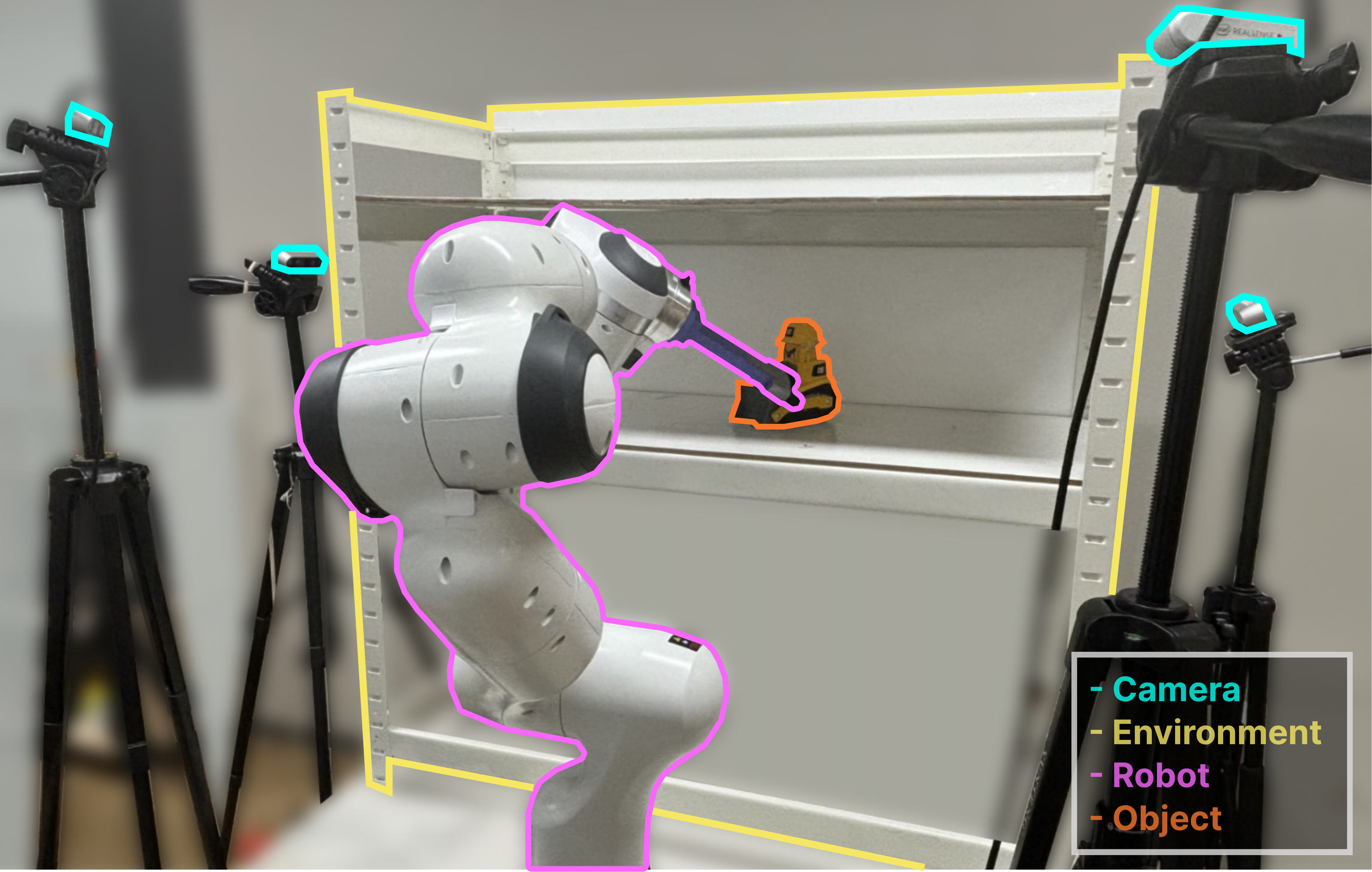}
        \caption{Example of our scene layout in the \emph{cabinet} domain.}
        \label{fig:realworld_drawer_setup}
    \end{subfigure}
    \hfill
    \begin{subfigure}[t]{0.49\linewidth}
        \includegraphics[width=\linewidth]{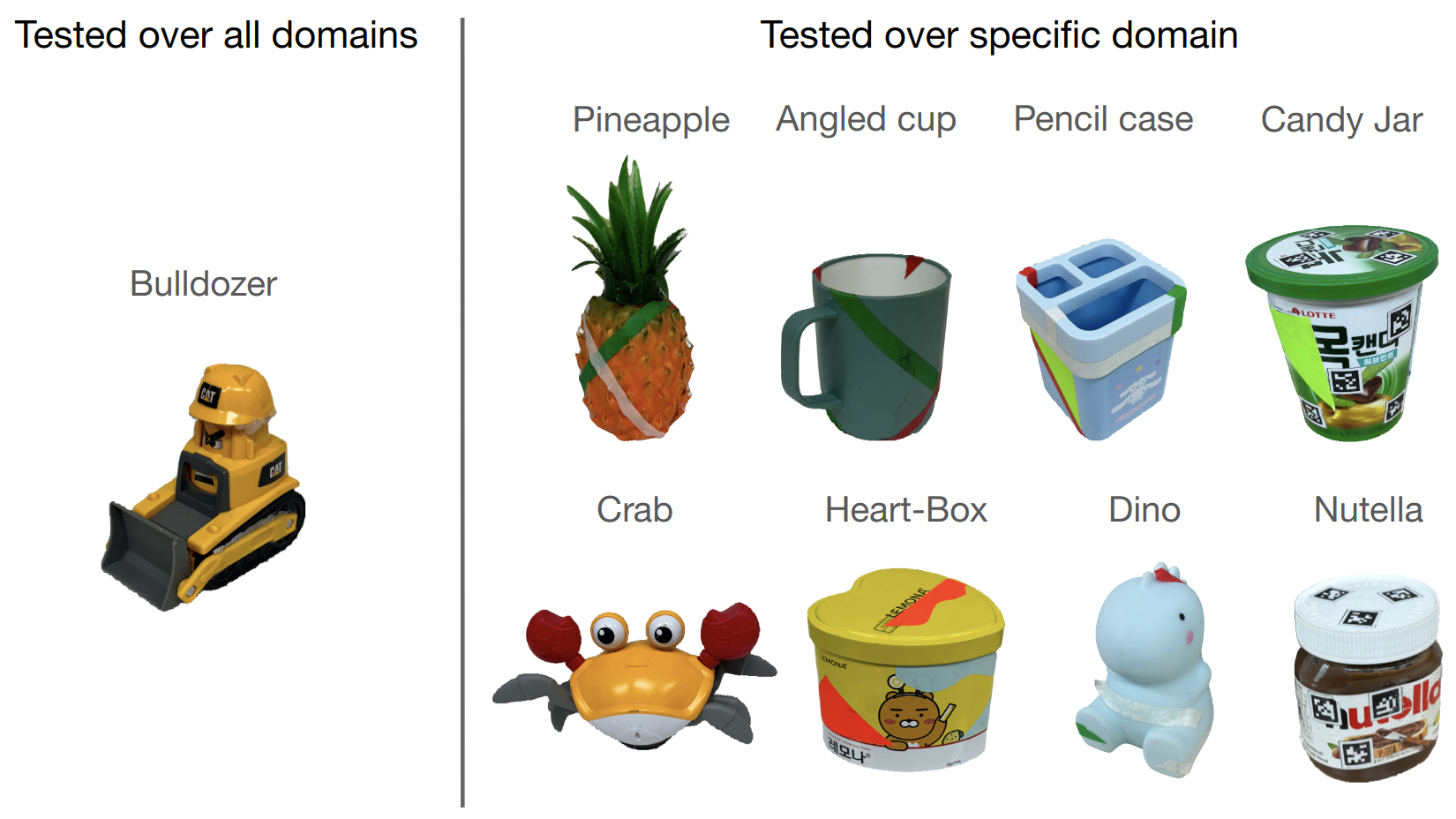}
        \caption{Our real-world objects, all unseen during policy training.}
        \label{fig:real-object}
    \end{subfigure}
    \caption{Our real-world experimental setup.}
    \label{fig:realworld-setup-both}
    \vspace{-3ex}
\end{figure}

To validate the real-world applicability and generalizability of our framework, we evaluate our policy in 9 real-world domains with novel everyday scenes and objects (Figure~\ref{fig:motivation}).
We test two objects in each domain: one object (a toy bulldozer), shared across all domains, and one random object (Figure~\ref{fig:real-object}),
each with five trials at different initial and goal poses.

For each scene, we mount four RealSense D435 cameras to observe the point clouds from multiple viewpoints, ensuring sufficient visibility of the object during execution (Figure~\ref{fig:realworld_drawer_setup}).
To distinguish the object cloud from the environment cloud, we use SAM~\cite{kirillov2023sam} to designate the initial object segmentation mask and utilize Cutie~\cite{cheng2024cutie} to track the object during manipulation. We use FoundationPose~\cite{wen2024foundationpose} to estimate the object's relative pose from the goal pose, using the view with the best visibility of the object (largest object segmentation mask) among the four cameras. We generate the environment point cloud by combining and filtering the point clouds from the depth cameras.
We replaced the robot's gripper to accommodate narrow environments, wrapped with a high-friction glove to reduce slipping. Further details on the real-world setup are in Appendix~\ref{sec:real-world-detail}.

Table~\ref{tab:real-result} shows the results of our policy across 9 real-world domains.
Overall, our policy demonstrates 78.9\% success rate,
indicating that our framework facilitates the policy to transfer to diverse, unseen real-world environments,
despite only training in a simulation.
The main failure modes of our policy are, in decreasing order of frequency:
torque limit violation (5.56\%); policy deadlock (4.44\%); dropping objects (4.44\%);
getting blocked by the environment (3.33\%); and perception error (3.33\%). Detailed descriptions of these failure modes are in Appendix~\ref{sec:failure-mode}.

\subsection{Emergence of skills in \textnormal{\oura{}}\label{sec:qualitative-analysis}}
\begin{figure*}[th!]
    \centering
    \includegraphics[width=0.85\linewidth]{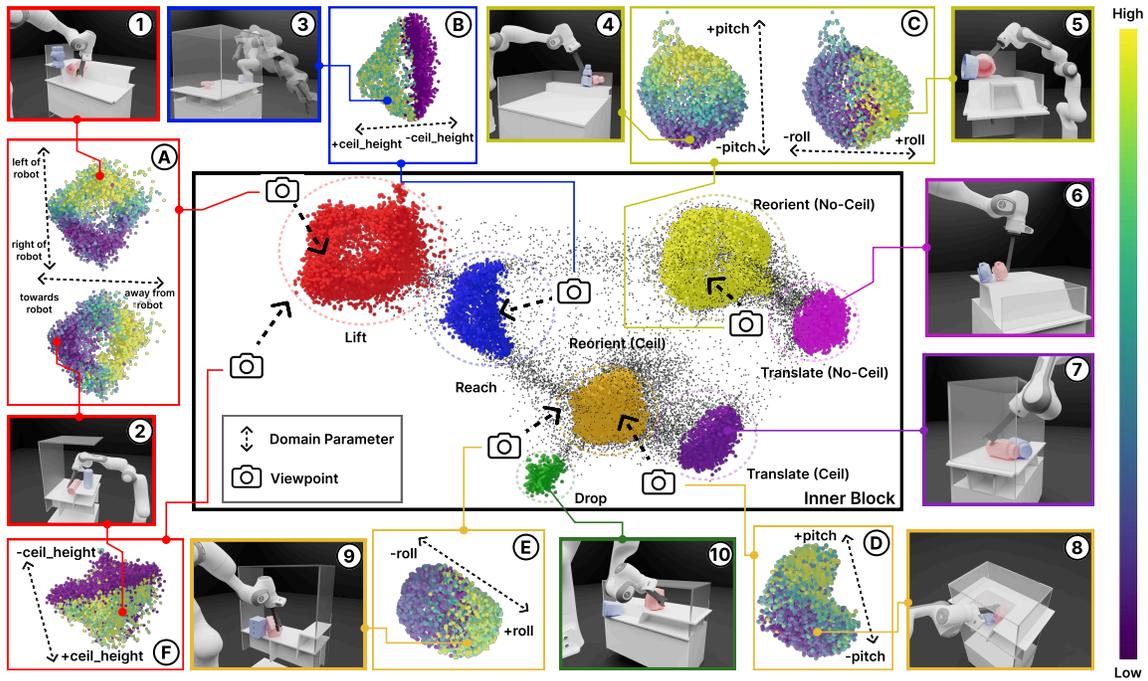}
    \caption{
    (inner block) UMAP projection of the modulation embedding $z_\tau$, colored by clusters from HDBScan. Unclustered points are in black.
    (outer block) Isolated view of each cluster, colored by a representative domain parameter, such as ceiling height or goal direction (to the left or right). The camera icon per each box denotes the viewpoint. The rendered scene shows a domain and state that generated an embedding in a cluster, with the red and blue objects indicating the current object pose and goal pose respectively.
    }
    \label{fig:qual-umap}
    \vspace{-2ex}
\end{figure*}

We show that \oura{} automatically discovers different manipulation skills and learns to sequence them. To do this, we inspect the modulation embedding $z_\tau$ (see Figure~\ref{fig:overall-arch}), which decides the activation weight of each module. We collect a dataset of $z_\tau$ by running a trained policy in 25,000 randomly sampled episodes in simulation. Since the high-dimensional $z_\tau$ is hard to interpret, we project $z_\tau$ into a three-dimensional manifold using UMAP~\cite{mcinnes2018umap} to visualize its structure.

To show that \oura{} discovers different skills, we apply HDBSCAN~\cite{mcinnes2017hdbscan} to these projections of $z_\tau$. Figure~\ref{fig:qual-umap} shows the result.
The \emph{inner block} of Figure~\ref{fig:qual-umap} shows the extracted clusters with different colors. We find that, without any manually designed bias or knowledge, these clusters naturally emerge and have semantically interpretable behaviors, such as
\emph{lifting} (red), \emph{reaching} (blue), \emph{reorienting} with (yellow) and without a ceiling (bright yellow), \emph{translation} with (purple) and without a ceiling (bright purple), and \emph{dropping} objects (green).

The \emph{outer blocks} marked with numbers show the rendering of the situations in which these embeddings have been used. They show that our policy also learns \textit{when} to use these skills based on the geometric constraint imposed by the environment, and the subgoal the robot is trying to achieve. For instance, to \emph{lift} objects over platforms (Figure~\ref{fig:qual-umap}, {\large \textcircled{\small 1}}), the policy must actively maintain contact between the object, wall, and the gripper. When \emph{dropping} objects (Figure~\ref{fig:qual-umap}, {\large \textcircled{\small 10}}), the robot must carefully prevent them from bouncing or rolling off the table. Similarly, the ceiling affects the policy’s \emph{reaching} strategies: with the ceiling above the object, the robot must approach the object laterally (Figure~\ref{fig:qual-umap}, {\large \textcircled{\small 8}}); in open environments, the robot can take overhand postures (Figure~\ref{fig:qual-umap}, {\large \textcircled{\small 4}}) instead.

The \emph{outer blocks} marked with alphabets in Figure~\ref{fig:qual-umap} show that intra-cluster variation captures their behavioral variations within a skill. For example, as you move 
horizontally
within the \emph{lifting} cluster (Figure~\ref{fig:qual-umap}, {\large \textcircled{\small a}}) it models behavior that pulls the object towards or away from the robot {\large \textcircled{\small 2}}, while 
the vertical direction
maps to its left or right {\large \textcircled{\small 1}}. While subtler than the \emph{categorical} differences across distinct skills like \emph{lifting} and \emph{reaching}, the emergence of such \emph{intra-cluster} variations shows that \oura{} also learns to adjust the module activations to implement finer behavioral nuances.

\begin{figure}[t!]
    \centering
    \begin{subfigure}{1.0\linewidth}
        \centering
        \includegraphics[width=0.9\linewidth]{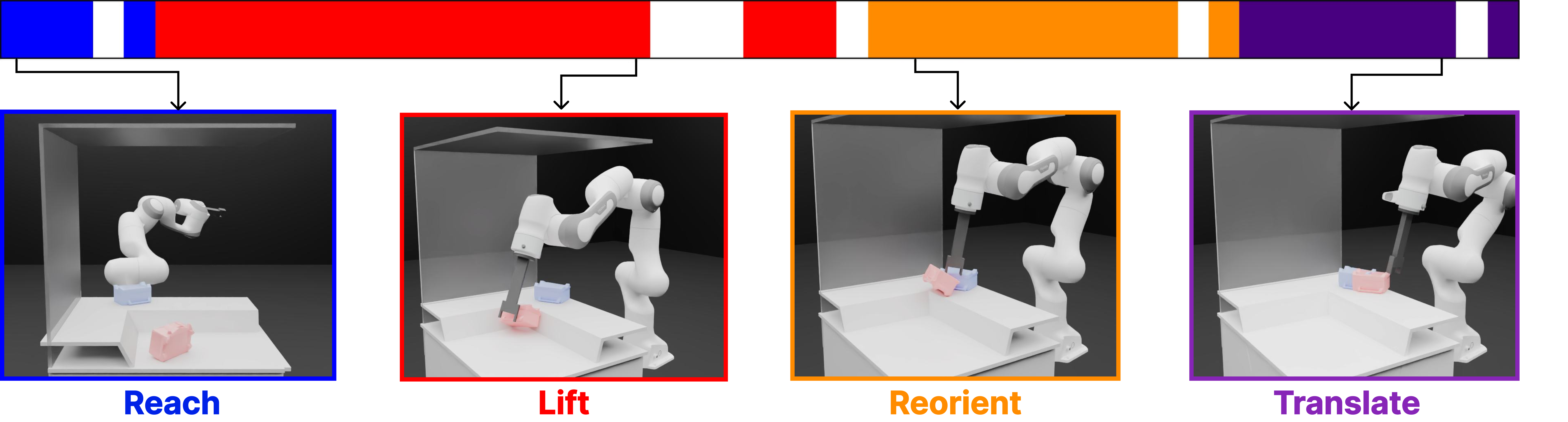}
        \caption{An episode where the agent lifts an object across a platform.}
        \label{fig:inter-phase-a}
    \end{subfigure}
    \begin{subfigure}{1.0\linewidth}
        \centering
        \includegraphics[width=0.9\linewidth]{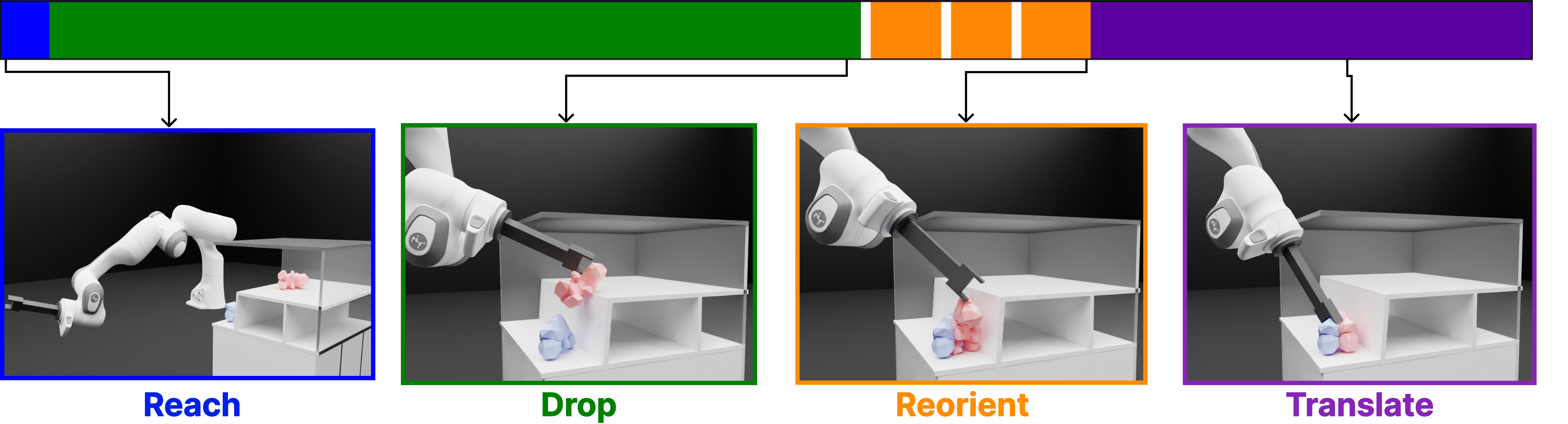}
        \caption{An episode where the agent drops an object to a lower elevation.}
        \label{fig:inter-phase-b}
    \end{subfigure}
    \caption{Illustration of how our architecture learns to use different skills.
    Color bar in each subfigure shows the cluster labels of $z_\tau$ at each step, and the bottom shows the domain rendering of representative keyframes. The red object is the current object pose, and the blue object is the goal pose.
    }
    \label{fig:inter-phase-route}
    \vspace{-3ex}
\end{figure}

To check if \oura{} can use these skills in sequence, we analyze how $z_\tau$ changes throughout a task.
To highlight the transitions, we label $z_{\tau}$ at each step of an episode based on the precomputed HDBScan clustering shown in Figure~\ref{fig:qual-umap}. Figure~\ref{fig:inter-phase-route} shows that the agent switches between behavioral clusters based on its internal subgoal: in Figure~\ref{fig:inter-phase-a}, the robot initiates with a \emph{reaching} skill to approach the object while avoiding obstacles. Afterward, the robot transitions to \emph{lift} the object to the top platform. After a successful lift, the robot \emph{reorients} the object to match the target orientation. Lastly, the robot \emph{translates} the object to its target pose. The sequence of $z_\tau$ changes when the problem changes: when the robot has to drop an object to a lower platform instead (Figure~\ref{fig:inter-phase-b}), the robot follows a different sequence (\emph{reach}-\emph{drop}-\emph{reorient}-\emph{translate}). This demonstrates that our architecture can (1) discover its own subgoals and (2) activate different modules to achieve different subgoals.

\subsection{Simulated Benchmark in Realistic Domains\label{sec:benchmark}}

\begin{figure}[t]
    \centering
    \includegraphics[width=0.85\linewidth]{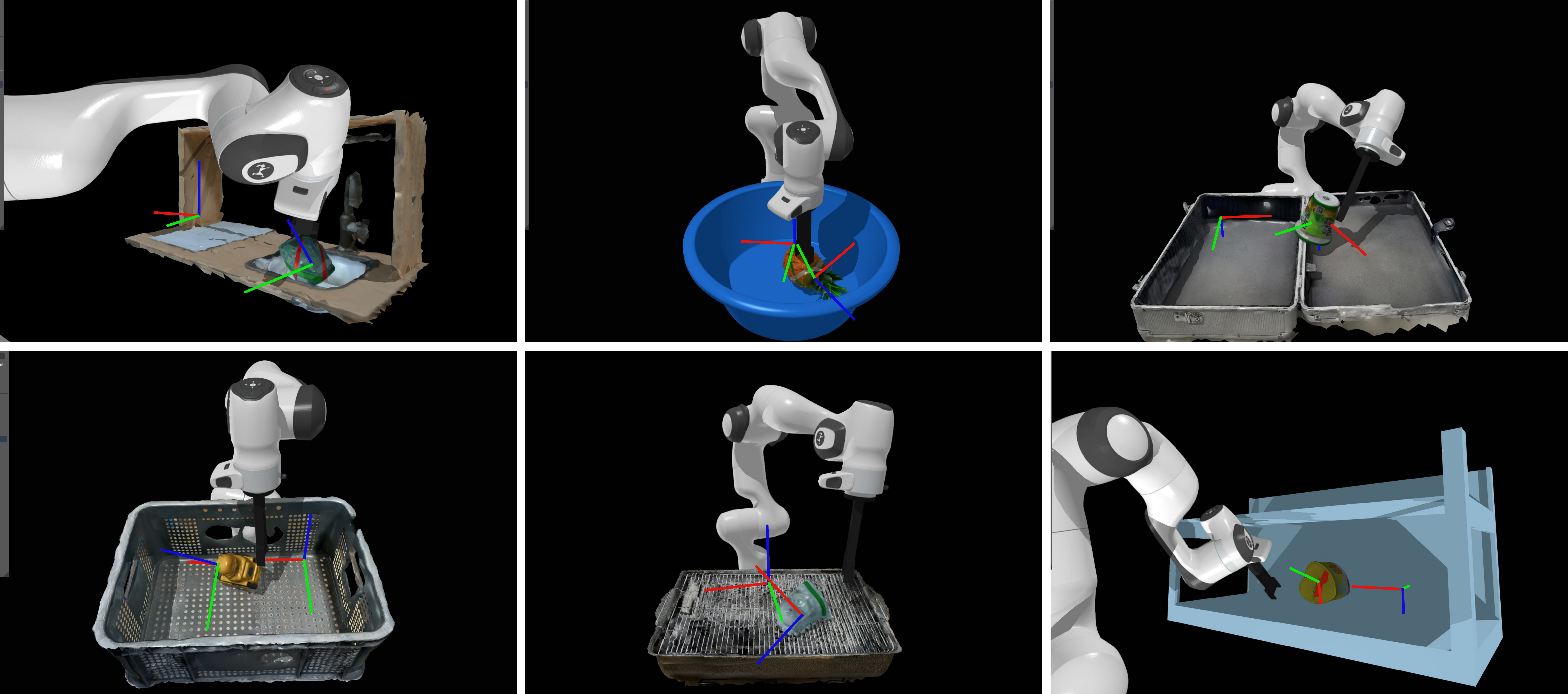}
    \caption{Sample environments in our simulated benchmark.
    From the top left: sink, circular bin, suitcase, basket, grill, and cabinet.
    The axis on each object indicates its current pose,
    while the other axis represents the target pose.}
    \label{fig:real2sim}
    \vspace{-2ex}
\end{figure}

We release a simulated digital twin of our nine real-world setups as a benchmark for non-prehensile manipulation (Figure~\ref{fig:real2sim}). 
The environment mesh is built using CAD, Nerfstudio~\cite{nerfstudio}, and Polycam. 
Our benchmark comprises 353 objects: 9 custom scans from the real world, 21 from GSO~\cite{laura2022GSO}, and 323 from DGN~\cite{wang2023dexgraspnet}. 
For each domain and object pair, we sample 5 stable initial- and goal-poses and 128 random collision-free robot initializations to evaluate generality.
Appendix~\ref{sec:r2s_detail}
details domain configurations and provides baseline results.

\section{Conclusion\label{sec:conclusion}}

In this work, we propose a novel framework
for non-prehensile manipulation in general environments
via deep reinforcement learning in a simulation.
Our framework comprises a modular architecture (\oura{}),
a contact-based object and environment representation (\ourr{}), and
a procedural domain generation algorithm for diverse environment geometries.
Compared to conventional architectures and standard representations, our framework
facilitates data- and time-efficient training
of a policy that generalizes to diverse and unseen scenes.
Despite solely training in synthetic environments in a simulation,
our policy zero-shot generalizes to unseen real-world environments and objects.
Overall, our combined framework achieves state-of-the-art performance in non-prehensile manipulation
of general objects in general environments.

\subsection{Limitations\label{sec:limit}}

Despite promising results, our approach has several limitations that can be addressed in future work:

\textbf{Improved efficiency of \oura{}.}
In \oura{}, the parameters of the base network are updated at every step. 
However, our qualitative analysis (Section~\ref{sec:qualitative-analysis}) shows that $z_\tau$, and the
resulting predictions for the module activation weights,
remain stable for prolonged periods until a transition triggers a change, such as a successful lift.
Thus, reusing the predicted network parameters over multiple steps can potentially reduce the computational overhead.

\textbf{Dynamics-aware object representation.} 
While \ourr{} effectively represents object and environment geometries, it neglects dynamic properties like mass and inertia,
which are critical for maneuvering objects with unusual dynamics, like roly-poly toys with non-uniform mass distributions.
As such, one intriguing future research direction is to extend the representation to include dynamics information,
potentially by incorporating memory~\cite{cho2014gru,gu2024mamba}.

\textbf{Generating fine-grained environment features.}
Our procedural generation pipeline relies on cuboidal primitives, limiting the diversity of
fine-grained geometric features (e.g., textures, curvatures, small overhangs).
While our experiments in the \textit{Grill}, \textit{Drawer}, and \textit{Circular Bin} environments
show that the policy can still adapt to uneven and curved surfaces,
diversifying procedural generation through approaches like geometric generative models~\cite{Siddiqui2024meshgpt},
may enhance the policy's environmental generalization capability.

\section*{Acknowledgements}
This work was supported by Institute of Information \& communications Technology Planning \& Evaluation (IITP) grant and National Research Foundation of Korea (NRF) funded by the Korea government(MSIT) (No.2019-0-00075, Artificial Intelligence Graduate School Program(KAIST)), (No.2022-0-00311, Development of Goal-Oriented Reinforcement Learning Techniques for Contact-Rich Robotic Manipulation of Everyday Objects), (No. 2022-0-00612, Geometric and Physical Commonsense Reasoning based Behavior Intelligence for Embodied AI), (No. RS-2024-00359085, Foundation model for learning-based humanoid robot that can understand and achieve language commands in unstructured human environments), (No. RS-2024-00509279, Global AI Frontier Lab).

\bibliographystyle{plainnat}
\bibliography{00main}
\newpage
\appendix

\subsection{Details on training \textnormal{\ourr{}} \label{sec:data-gen-unicorn}}
\subsubsection{Contact dataset generation}
Since predicting contacts between distant or fully overlapping objects is trivial, we prioritize sampling \emph{near-contact} object configurations
to encourage our model to learn informative representations.
To achieve this, we follow CORN~\cite{cho2024corn} and randomly sample two objects from the object dataset and position each object at a random SE(3) pose.
Since this initial placement is unlikely to result in objects in colliding configurations,
we first move the objects tangent to each other by measuring the shortest displacement between
the two objects, and translating one of the objects by that amount
so that they come into contact with each other.
Afterward, we apply a small Gaussian noise to the poses
so that the objects either slightly clip into each other or narrowly remain collision-free.

After positioning the objects, we generate the point clouds and contact labels. We sample the point clouds uniformly from the surface of each object,
then label the points based on whether they fall within the other object. Since computing point-mesh intersection is often unreliable, we first apply convex decomposition to each object, then compute whether each point falls within any of the other object's convex parts, and vice versa.
Overall, this procedure yields approximately half of the dataset comprising objects in colliding configurations,
with the other half in near-contact configurations; the representation model must learn to distinguish the two scenarios. We iterate this procedure to generate a dataset comprising 500,000 point cloud pairs and their contact labels.

\subsubsection{Details on pretraining pipeline\label{sec:pretrain-arch-details}}
The contact decoder is a three-layer conditional MLP (CMLP), where each layer is a residual block with conditional batch normalization (CBN)~\cite{mescheder2019occupancy}. CBN transforms each layer's output features by applying batch normalization, whose affine parameters are mapped from the conditioning input $z_N^{B}$ with a single linear layer.

During training, we apply data augmentation by rotating, translating, and scaling both clouds, plus a small Gaussian noise. After patchifying the point clouds, we adjust the proportion of the inputs to the contact decoder so that approximately half of the input pairs are in contact (positive). This is done by resampling the positive patches with probability $f/P$ and negative patches with probability $(1-f)/N$, where $f$ denotes the target fraction of positive labels, $P$ denotes the number of positive patches, and $N$ denotes the number of negative patches. The hyperparameters for pretraining are in Table~\ref{tab:pretrain-params}.

\begin{table}[thbp]
\centering
\caption{Pretraining pipeline hyperparameters.}
\label{tab:pretrain-params}
\resizebox{0.75\columnwidth}{!}{%
\begin{tabular}{@{}ll@{}}
\toprule
\textbf{Hyperparameter}    & \textbf{Value}          \\ \midrule
Batch size                 & 1024                    \\
Optimizer                  & SAM~\cite{foret2020sam} \\
Learning rate schedule     & cosine                  \\
Base learning rate         & 0.0002                  \\
Min. learning rate         & 1e-6                    \\
Max. gradient norm         & 1,000                   \\
Weight decay               & 0.001                   \\
Rotational augmentation    & ($-\pi$, $+\pi$)        \\
Translational augmentation & (-0.1, +0.1)            \\
Scale augmentation         & $(e^{-1}, e^{+1})$      \\
Noise augmentation         & 0.01                    \\
Positive patch fraction    & 0.5                     \\
Decoder size        & (128, 128)              \\ \bottomrule
\end{tabular}%
}
\end{table}

\subsection{Details on Procedural Domain Generation\label{sec:domain-gen-detail}}

\subsubsection{Environment Geometry\label{sec:proc-gen-env-geom}}

\begin{figure*}[thbp]
    \centering
    \includegraphics[width=0.8\linewidth]{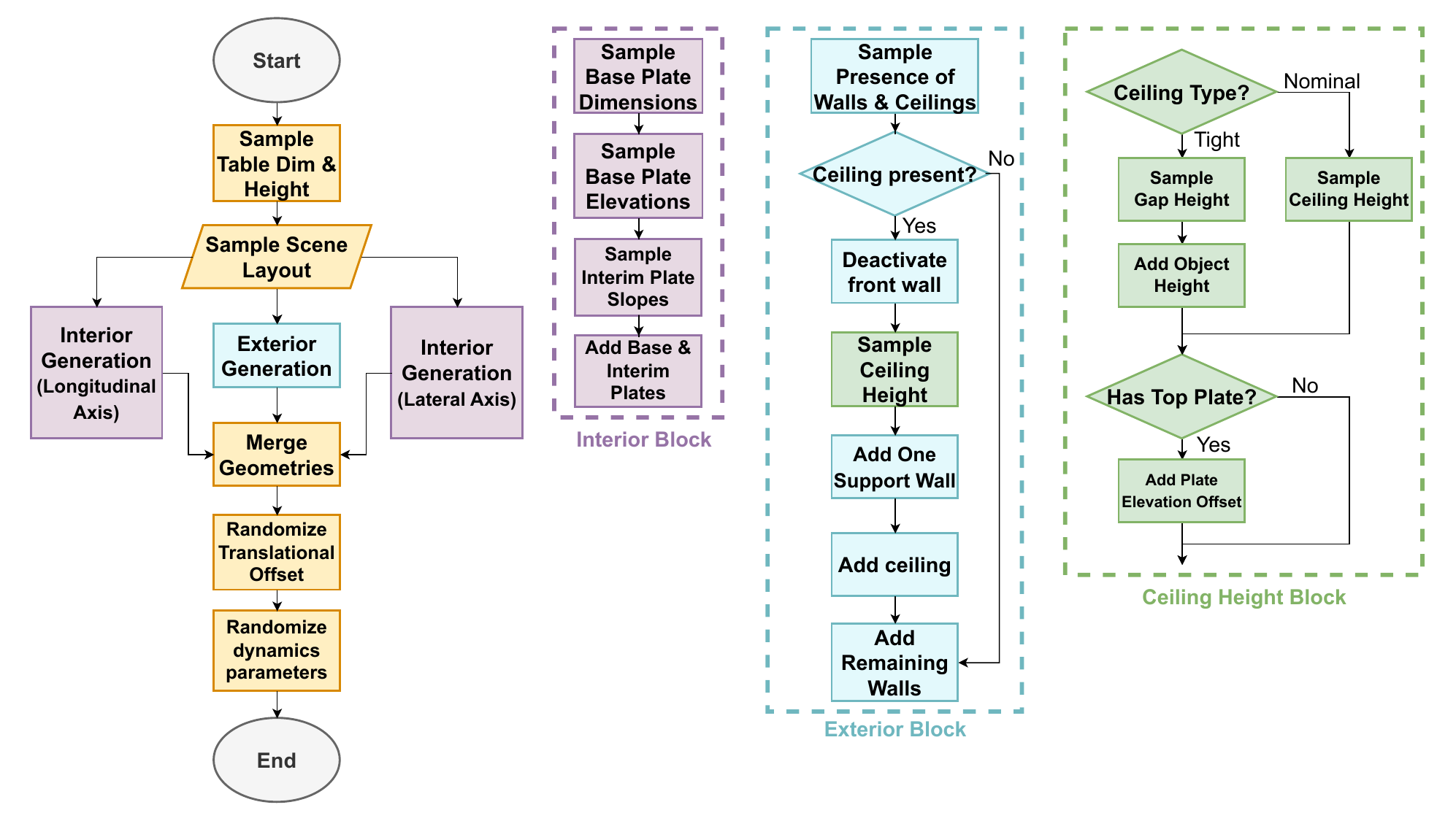}
    \caption{Flowchart for our procedural generation pipeline. The leftmost flowchart describes the overall procedure; the expanded blocks (purple, cyan, and green) on the right show the subroutines in detail.}
    \label{fig:proc-gen-flow}
\end{figure*}

The flowchart in Figure~\ref{fig:proc-gen-flow} depicts our procedural generation pipeline, which samples environment geometries by composing a set of cuboidal primitives. Our pipeline is split into two main parts: the \emph{interior} of the domain (Figure~\ref{fig:proc-gen-flow}, purple),
comprising the planar surface at various elevations and sloped hills formed by the \emph{lateral} and \emph{longitudinal} base plates (see Figure~\ref{fig:proc-gen-schem}),
and the \emph{exterior} of the domain (Figure~\ref{fig:proc-gen-flow}, cyan) comprising the ceiling and walls.

Our procedure begins by sampling the overall scene dimensions, i.e., whether a given table will be narrow or wide. This determines the overall scale of the cuboid primitives to fit within the bounds of the scene dimensions. Based on this, we compose the \emph{interior} of the domain from a set of cuboidal plates. Each of the lateral and longitudinal axes comprises five plates: three \emph{planar} plates form level surfaces at various heights, joined by two \emph{interim} plates that form walls at various slopes.

\begin{figure}[thbp]
\centering
\includegraphics[width=1.0\linewidth]{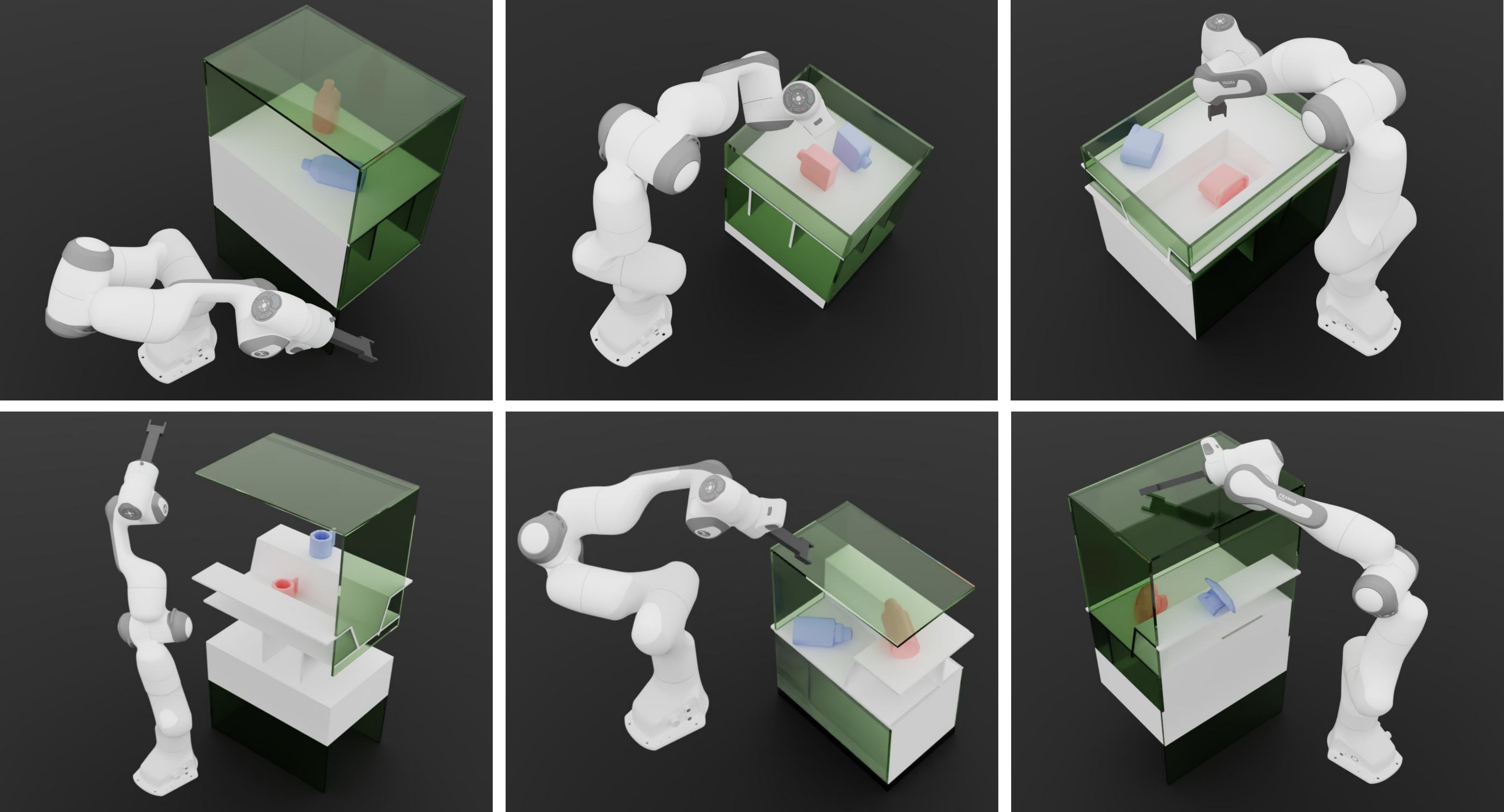}
         \caption{Additional examples from our environment generation algorithm. While fully procedural and randomized, our pipeline yields geometries resembling those of real-world scenes, such as (from the top left): cabinet, basket, sink, valley, countertop, and step. Walls and ceilings are rendered with transparent green to distinguish them from the base plates. The red object indicates the object at initialization, and the blue object indicates the goal pose of the object.
         }
         \label{fig:proc-gen-sim}
        \vspace{-3ex}
\end{figure}

To generate the \emph{interior} of the domain (Figure~\ref{fig:proc-gen-flow}, purple), we start by randomly sampling the plate dimensions,
while ensuring the sum of the \emph{planar} plate dimensions along each axis does not exceed the scene bounds.
Afterward, the elevations of planar plates are randomly sampled by designating whether each plate is a \emph{top} plate or a \emph{bottom} plate,
separated by a randomly sampled difference in elevation.
Afterward, we sample the angles of the slopes corresponding to the interim plates that connect between two planar plates.
Combining the planar and interim plates for both axes constructs the base surface of the domain,
resulting in structural layouts such as sinks, bumps, or valleys (Figure~\ref{fig:proc-gen-sim}). 

Next, we generate the \emph{exterior} of the domain, composed of walls and ceilings (Figure~\ref{fig:proc-gen-flow}, cyan). Since these obstructions directly impact the accessibility of the workspace, we take a multi-step approach for positioning them. First, we randomly sample whether the ceiling should exist in the domain. Then, if the ceiling is present, the front-facing wall is deactivated to allow for the robot's entry. The remaining walls are randomly configured while ensuring at least one load-bearing wall is added on one of the four sides to support the ceiling. The disabled features are hidden beneath the tabletop to prevent interaction with other simulated entities.

Since the height of the ceiling heavily influences the accessibility of the object,
we implement two different procedures to determine ceiling heights: \emph{nominal} and \emph{tight} (Figure~\ref{fig:proc-gen-flow}, green). The heights of \emph{nominal}-type ceilings are simply sampled from a uniform distribution with a sufficient margin.
On the other hand, heights of \emph{tight}-type ceilings are determined by adding a small \emph{gap} to the object height,
where the height of the gap is drawn from a uniform distribution.
Lastly, in both cases, if the interior of the domain has elevated platforms such as steps or bumps,
we raise the ceiling heights by the height of the platform to prevent the object from intersecting with the ceiling.

After constructing the overall environment geometry, we further randomize the environment by shifting its position along each of the x, y, and z axes and adjusting its surface friction coefficients.
Table~\ref {tab:env-param} lists the parameters for our procedural generation scheme.

\begin{table}[htbp]
\centering
\caption{Scene parameters and their ranges in our procedural generation pipeline. All angles are in degrees, and dimensions are in meters. \textsc{U}: uniform distribution; \textsc{B}: Bernoulli distribution.}
\label{tab:env-param}
\resizebox{0.6\columnwidth}{!}{%
\begin{tabular}{@{}ll@{}}
\toprule
\textbf{Parameter} & \textbf{Value} \\ \midrule
table\_dim.x & U(0.255, 0.51) \\
table\_dim.y & U(0.325, 0.65) \\
table\_dim.z & U(0.2, 0.4) \\
table\_pos.x & U(0.0, 0.1) \\
table\_pos.y & U(-0.15, 0.15) \\
table\_pos.z & U(0.1, 0.8) \\
ramp\_angle & U(0.0, 30) $\times$ 4 \\
plate\_height & U(0.0, 0.15) $\times$ 6 \\
ceiling\_height & U(0.3, 0.5) \\
gap\_height & U(0.03, 0.05) \\
ceil\_mask & B(0.5) \\
wall\_mask & B(0.5) $\times$ 4 \\
table\_friction & U(0.2, 0.6) \\ \bottomrule
\end{tabular}%
}
\end{table}

\subsubsection{Object Placement\label{sec:proc-gen-place}}

Since our environment geometries change at the start of each episode, we must sample stable and collision-free object placements \emph{online} to compute the initial and goal poses.
Since this process is time-consuming, we precompute a set of stable object orientations as in \cite{cho2024corn} by dropping them in a simulation.
Afterward, we also precompute the planar radius for each of the object's stable orientations as the distance to the farthest point on the object from the object's center.

For each episode, we sample object poses by combining one of the stable orientations with the position sampled from the horizontal plates in the environment. To compute collision-free and stable placements, we use the object's precomputed radius to serve as the minimum distance away from the nearest wall or edge of the plates.

We first sample the goal poses for the objects, then sample their initial poses while ensuring sufficient separation from the goal in terms of both its position and orientation. This prevents the episode from terminating in success immediately. When sampling initial and goal poses in domains with height differences between plates, we bias the proportions to encourage the goal poses to be on elevated platforms compared to the initial poses, which favors sampling more challenging tasks where the robot must lift objects across a slope or a wall in the environment.

\subsection{Simulation-to-Real Transfer Pipeline\label{sec:sim2real-transfer}}
The policy trained in simulation cannot be directly transferred to the real world.
This is for two reasons. First, in contact-rich scenarios like non-prehensile object manipulation,
the frequent contact between the robot and the object or the environment is prone to trigger hardware torque-limit violations. Second, the real-world policy does not have access to privileged information as in the simulation, such as the object's mass and dynamics properties.

To overcome the torque-limit violations, we adopt two main strategies: \emph{action magnitude curriculum} and
\emph{cartesian-space action clipping}, both of which reduce the scale of the policy actions to encourage conservative motions. To overcome the second issue of unavailable observations, we adopt \emph{teacher-student distillation}, in which the student replicates the teacher's actions solely from observations that would be available in the real world. 

\subsubsection{Action Magnitude Curriculum\label{sec:s2r-action-mag}}
In our domain, the robot frequently experiences contact with the object or the environment.
However, when the policy operates at high velocity under mismatched robot dynamics due to the sim-to-real gap,
such frequent contact may trigger hardware torque-limit violation when the impact is larger than expected. 

As the robot must move the object through contact, the impact is inevitable. The policy cannot readily learn to prevent high-impact collisions either,
since accurately reproducing the exact impact force in the simulation is
challenging due to the sim-to-real gap.
To circumvent this, we encourage the policy to perform generally conservative motions.

To this end, we adopt an action-magnitude curriculum inspired by the scheme from Kim et al.~\cite{kim2023crm}, where the maximum bounds of the subgoal residual that the policy can output is reduced gradually. During initial training,
we start with the maximum joint-space residual magnitude $\xi=\xi_{max}$ to facilitate the policy's exploration. We then gradually reduce $\xi$ to the target magnitude $\xi^{*}$, deemed safer for execution on the real robot. We apply different $\xi_{max}$ and $\xi^{*}$ values for the large and small joints for the FR3 arm. Our reduction schedule follows a geometric sequence with ratio $\frac{\xi^{*}}{\xi_{max}}^{\frac{N_s}{N_t}}$, where $N_t$ and $N_s$ are hyperparameters that denote the total number of simulation steps for annealing and the interval between successive annealing steps, respectively.
Detailed hyperparameters are in Table~\ref{tab:sim2real-params}

\begin{table}[thbp]
\centering
\caption{Sim2real hyperparameters.}
\label{tab:sim2real-params}
\resizebox{0.5\columnwidth}{!}{%
\begin{tabular}{@{}cc@{}}
\toprule
\multicolumn{1}{l}{\textbf{Hyperparameter}} & \multicolumn{1}{l}{\textbf{Value}} \\ \midrule
$\xi^{*}$ (large joint)                     & 0.16                               \\
$\xi_{\text{max}}$ (large joint)            & 0.26                               \\
$\xi^{*}$ (small joint)                     & 0.08                               \\
$\xi_{\text{max}}$ (small joint)            & 0.21                               \\
$N_s$                                       & 1024                               \\
$N_t$                                       & 2e6                                \\
$\epsilon_x$                                & 0.12                               \\
$\epsilon_x^{\text{max}}$                   & 0.24                               \\
$\alpha$                                    & 0.8                                \\ \bottomrule
\end{tabular}%
}
\end{table}
\subsubsection{Cartesian-space Action Clipping\label{sec:s2r-cart-clip}}
When using joint-space actions, multiple joints can simultaneously contribute to the same Cartesian direction.
Despite reducing overall action magnitude,
the sum of individual joint actions
may still result in large end-effector movements,
leading to high-force impacts that abort the robot.
While further reducing the joint space could mitigate this,
it would significantly degrade the robot's dexterity and strength.

\begin{algorithm}[ht]
    \caption{Cartesian-space action clipping algorithm.}
    \label{eq:cclip}
    \begin{algorithmic}[1]
    \Require Policy $\pi_\theta$, robot joint position $q$, cartesian action bound $\epsilon_x$, damping parameter $\lambda$
    \Ensure Clamped joint-space subgoal residual $\Delta q_{clamped}$
    \State $J = \frac{\delta x}{\delta q} |_q$
    \State $\Delta q \sim \pi_\theta(s)$
    \State $\Delta x = J \Delta q$
    \State $\Delta x_{excess}$ = $\Delta x \cdot \frac{||\Delta x||-\epsilon_x}{||\Delta x||}$
    \State $\Delta q_{excess}$ = $J^T(JJ^T+\lambda^2I)^{-1}\Delta x_{excess}$
    \State $\Delta q_{clamped}$ = $\Delta q - \Delta q_{excess}$
    \end{algorithmic}
\end{algorithm}

Instead, we devise a scheme to clamp the joint residuals based on the projected end-effector space movement to reside within the bound $\epsilon_x$ with minimal change to the original action, shown in Algorithm~\ref{eq:cclip}. We first compute the Jacobian $J$ of the robot in the current configuration (line 1). Then, we project the joint-space residuals $\Delta q$ to the Cartesian-space end-effector movement $\Delta x$ based on the Jacobian $J$ (line 2-3). If the estimated end-effector movement exceeds the predefined Cartesian bound $\epsilon_x$, we compute the excess movement $\Delta x_{excess}$ compared to $\epsilon_x$ (line 4), then subtract the excess from the original joint residuals by re-projecting $\Delta x_{excess}$ to $\Delta q_{excess}$ based on damped least-squares method (line 5-6). Since clipping the action bounds potentially subjects the policy to a large behavioral change,
we fine-tune the policy by starting with a large Cartesian bound $\epsilon_x^{\text{max}}$ and gradually annealing it down to the target $\epsilon_x$.

In addition to the magnitude scaling and Cartesian space clipping, we introduce a joint residual smoothing and energy-reducing loss
for suppressing jerky motion during real-world deployment.
Specifically, the smoothed residual $\Delta\Bar{q}_t$ is computed from the original joint residual $\Delta q_t$ with exponential moving average: $\Delta\Bar{q}_t=\alpha q_t + (1-\alpha)\Delta \Bar{q}_{t-1}$, where $\alpha \in [0,1]$. The smoothed value $\Delta\Bar{q}_t$ is then used to control the robot. The energy-reducing loss, computed by L2 norm of power $||\sum_{i=1}^{7}\tau_i\Dot{q}_i||_2$, is added as a regularizing loss during policy training.

\begin{figure}[tb]
\centering
    \includegraphics[width=\linewidth]{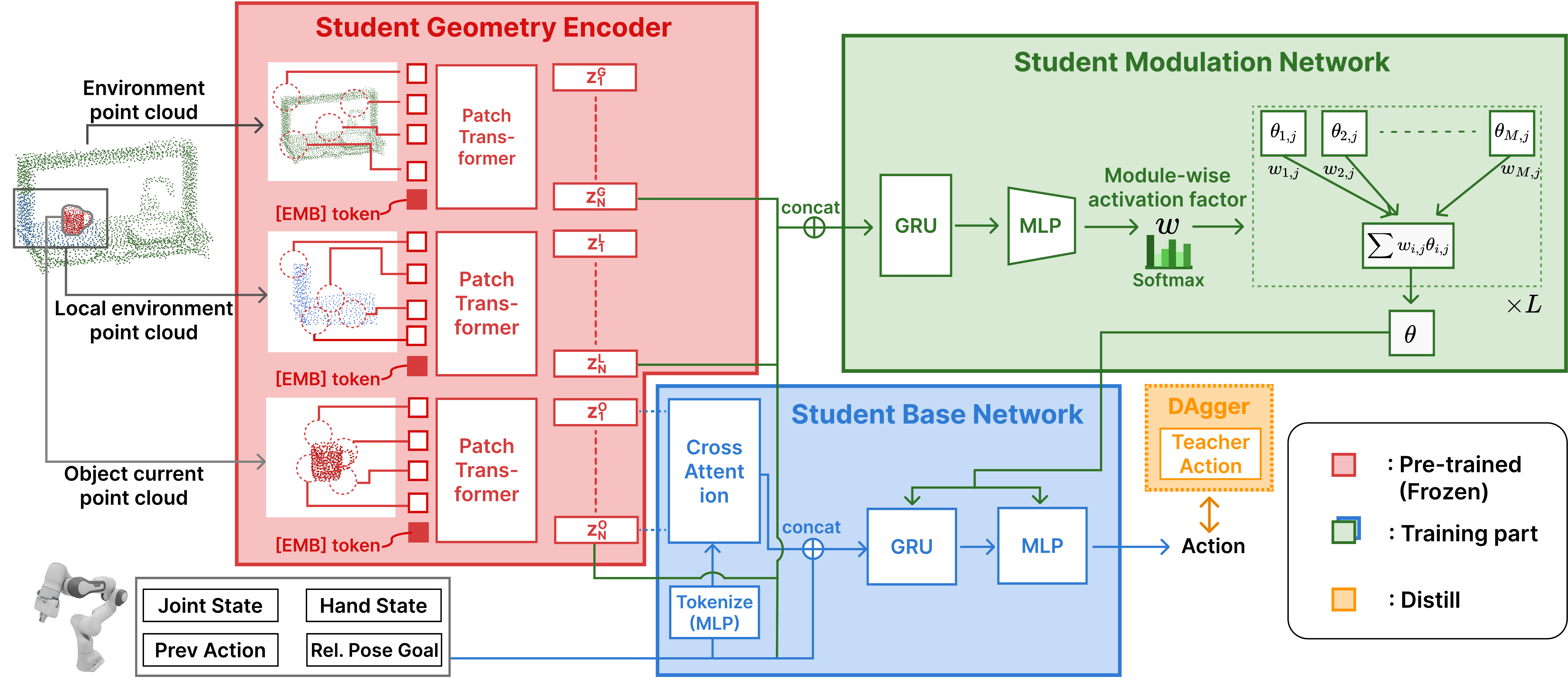}
    \caption{Illustration of student policy architecture. As in the teacher network, the student architecture comprises a geometry encoder, a modulation network, and a base network.}
    \label{fig:student-policy-arch}
\end{figure}
\subsubsection{Teacher-Student Distillation\label{sec:s2r-distill}} During policy training, we utilize privileged information such as coefficients of friction, restitution, and the object's inertial parameters and object velocity, which are not generally observable in the real world. As a result, the trained policy cannot operate without these quantities.
To address this, we distill the trained \emph{teacher} policy into a \emph{student} policy that operates based on observations available in the real world. During distillation, we employ DAgger~\cite{ross2011dagger}, where the student policy replicates the actions of the teacher solely based on the available observations through supervised learning during the simulation rollout. 

As illustrated in Figure~\ref{fig:student-policy-arch}, the student policy shares a similar architecture with the teacher policy, consisting of a geometry encoder (red), a modulation network (green), and a base network (blue). As in the teacher, the student modulation network generates the weights for the student base network based on the outputs of the geometry encoder. The student base network then produces actions based on both the generated weights from the modulation network and the state inputs.

However, because the student policy must estimate the teacher policy's actions with limited information, we incorporate a Gated Recurrent Unit (GRU)~\citep{cho2014gru} into both the modulation network and the base network. This allows the student model to aggregate information from previous observations, helping it infer the teacher policy's actions more effectively.

\subsection{Details on MDP Design\label{sec:mdp-design-extra}} 

\begin{table*}[ht]
\centering
\caption{Summary of our MDP, in terms of state, action, and reward components. Each component is denoted by its name, shorthand symbol, dimensionality and a brief description. $\dag$: only used in simulation.}
\label{tab:mdp_summary}
\centering
\label{tab:state_components}
\resizebox{0.9\linewidth}{!}{%
\begin{tabular}{|c|c|c|c|}
\hline
\textbf{State Component}        & \textbf{Symbol} & \textbf{Dimension} & \textbf{Description}                                  \\ \hline
$\text{Object state}^{\dag}$       & $x_t^o$ & $\mathbb{R}^{15}$ & Object pose and velocity                 \\ \hline
Robot state          & $x_t^q$         & $\mathbb{R}^{14}$  & Joint positions and velocities                        \\ \hline
End-effector pose    & $x_t^{EE}$      & $\mathbb{R}^9$     & Pose of the robot's end-effector                      \\ \hline
$\text{Physics parameters}^{\dag}$ & $\nu$   & $\mathbb{R}^6$    & Mass, friction, restitution of object and friction of robot and environment\\ \hline
Object geometry      & $G_o$           & $\mathbb{R}^{512\times 3}$ & Surface-sampled point cloud of the object             \\ \hline
Environment geometry & $G_e$           & $\mathbb{R}^{512\times 3}$ & Surface-sampled point cloud of the environment        \\ \hline
Goal pose            & $T_g$           & $\mathbb{R}^9$     & Target pose for the object, relative to current pose \\ \hline
\hline
\textbf{Action Component}        & \textbf{Symbol} & \textbf{Dimension} & \textbf{Description}                                  \\ \hline
Joint-space subgoal residuals & $\Delta q$ & $\mathbb{R}^7$ & Desired changes in joint positions \\ \hline
Proportional gains & $k_p$ & $\mathbb{R}^7$ & Joint-space proportional gains \\ \hline
Damping factors & $\rho$ & $\mathbb{R}^7$ & Factors for computing damping terms \\ \hline
\hline
\textbf{Reward Component}        & \textbf{Symbol} & \textbf{Dimension} & \textbf{Description}                                  \\ \hline
Task success reward & $r_{\text{s}}$ & $\mathbb{R}^{1}$ & Reward for task success \\ \hline
Goal-reaching reward & $r_{\text{r}}$ & $\mathbb{R}^{1}$ & Reward for moving object towards goal \\ \hline
Contact-inducing reward & $r_{\text{c}}$ & $\mathbb{R}^{1}$ & Reward for moving gripper towards object \\ \hline
\end{tabular}%
}
\end{table*}

\begin{table}[htb]
\centering
\caption{Hyperparmeters for the reward terms.}
\label{tab:reward-params}
\resizebox{0.95\linewidth}{!}{%
\begin{tabular}{@{}ccc@{}}
\toprule
\textbf{Parameter} & \textbf{Value} & \textbf{Description} \\ \midrule
$\lambda_r$ & 0.15 & Goal-reaching reward coefficient \\
$\lambda_c$ & 0.03 & Contact-inducing reward coefficient \\
$c_g$ & 3.0 & Scale for goal-reaching distance potential \\
$c_r$ & 3.0 & Scale for contact-inducing distance potential \\ \bottomrule
\end{tabular}%
}
\end{table}
The state, action, and reward components of MDP are summarized in Table~\ref{tab:mdp_summary}. 
Hyperparameters of the reward terms are shown in Table~\ref{tab:reward-params}.
\subsection{Point cloud sampling process\label{sec:cloud-sample}}
Our policy takes three types of point-cloud inputs: \emph{global} scene cloud, \emph{local} scene cloud, and \emph{object} clouds.
To obtain these inputs, we need to sample the points from the underlying meshes.
For the \emph{object} cloud, we can pre-sample the point clouds from the underlying mesh,
then transform its point cloud to the current pose.
However, obtaining environmental point clouds is non-trivial:
since our scene is constructed dynamically, the corresponding point cloud must change across episodes.
Thus, the point clouds cannot be pre-sampled, and an efficient online sampling procedure is necessary.

To sample the surface point clouds from a union of primitives,
we must determine the subsection of the surfaces that form the exterior of the composed cuboids.
While the simplest solution is to compute the boolean union of environment meshes,
this operation is computationally costly as it cannot be parallelized.
Instead, we subdivide the cuboid surfaces into a set of non-intersecting triangles, then cull the triangles that are contained by the cuboids. Afterward, we sample the point clouds proportional to the area of the remaining non-occluded triangles.
By vectorizing this operation, we can efficiently sample the environmental point clouds by leveraging GPU-based acceleration.

Afterward, the \emph{local} point cloud is sampled by selecting the points on the global cloud nearest to the object. 
For computational efficiency, we pre-sample a set of 64 keypoints on the object surface via FPS,
then sub-sample 512 points from the global scene cloud with the lowest distance to the nearest object keypoint in the current pose.

\subsection{Details on Baseline Architectures\label{sec:baseline-arch}}

{\ourr{}\textsc{-SM}} uses \ourr{} as the input representation (Section III-B), combined with the architecture from Soft Modularization~\cite{yang2020softmodule}. 
The implementation is adapted from the author's original code to operate with Isaac Gym~\cite{isaacgym}. For a fair comparison, we use the same number of modules and base network size as in \oura{}. The module activations are embedded with 128 dimensions, and passed through a two-layer MLP with a hidden dimension of 128 for each layer of the base network.

{\textsc{PointGPT}-\oura{}} uses \oura{} with the PointGPT-S~\cite{chen2023pointgpt} representation model, utilizing the code and pre-trained weights released by the authors. However, since the original model is memory-intensive and computationally slow, we use TensorRT~\cite{TensorRT} to optimize the model.

{\ourr{}\textsc{-MONO}} is a standard MLP using wider hidden layers (with dimensions [768, 384, 384, 192]) to match the number of trainable parameters in \ourr{}-\oura{}. It uses Tanh activation and layer normalization in the hidden layers. When computing its input, we apply the same cross-attention as in \oura{} to the geometric embeddings, then concatenate the resulting embeddings with all non-geometric state inputs before feeding them into the network.

{\ourr{}\textsc{-HYPER}} uses a hypernetwork~\cite{ha2017hypernetworks} to output the parameters of the base network. As directly predicting all parameters requires an impossibly large hypernetwork, we instead design \ourr{}\textsc{-HYPER} to predict \emph{low-rank} decompositions of base network parameters. For each layer of the base network of form $x_{i} = \sigma(Wx_{i-1}+b)$ where $i$ denotes layer index, the hypernetwork $\phi$ outputs $\phi(z)=\{W_l,W_r,b\}$ and constructs $W=W_lW_r$ where $W \in \mathbb{R}^{N\times M}$, $W_l \in \mathbb{R}^{N\times k}$, and $W_r \in \mathbb{R}^{k \times M}$, where $k$ denotes the rank, thus reducing the output dimensions from $N\times M + M$ to $(N+M) \times k + M$. In all our experiments, we configure the rank to be 16, and the hypernetwork is an MLP with hidden dimensions $[256,256]$ using Tanh activation and LayerNorm in the interim layers.

{\ourr{}\textsc{-Transformer}} is a four-layer transformer, where each layer uses the embedding dimension of 512 and four attention heads. For computational efficiency, we leverage FlashAttention~\cite{dao2022flashattention} in our implementation.
The transformer receives learnable action and value tokens, geometric embeddings (generated by the same cross-attention used in \oura{}), and non-geometric state inputs tokenized with a linear layer.
After passing these tokens through the transformer layers, a two-layer MLP with a 256-dimensional hidden layer maps the action and value token embeddings to the robot’s action and state value, respectively.
Detailed hyperparameters for the network architectures and PPO training are described in Table~\ref{tab:network} and Table~\ref{tab:PPO}, respectively. 

\begin{table*}[thb]
\centering
\caption{Network Hyperparameters.}
\label{tab:network}
\begin{tabular}{cccccc}
\hline
\textbf{Hyperparameter} &
  \textbf{Value} &
  \textbf{Hyperparameter} &
  \textbf{Value} &
  \textbf{Hyperparameter} &
  \textbf{Value} \\ \hline
Num. points &
  512 &
  Num. encoder layers &
  4 &
  \begin{tabular}[c]{@{}c@{}}Modulation\\ Network\end{tabular} &
  MLP (256, 256) \\
Num. patches &
  16 &
  Num. self-attn heads &
  4 &
  Actor &
  MLP (256, 128, 128, 64) \\
Patch size &
  32 &
  Cross-attn embedding dim. &
  64 &
  Critic &
  MLP (256, 128, 128, 64) \\
Embedding dim. &
  128 &
  \begin{tabular}[c]{@{}c@{}}Num. cross-attn heads \\ (object / others)\end{tabular} &
  8/4 &
  \begin{tabular}[c]{@{}c@{}}Num. modules\end{tabular} &
  8 \\ \bottomrule
\end{tabular}
\end{table*}

\begin{table}[thbp]
\centering
\caption{PPO Hyperparameters.}
\label{tab:PPO}
\resizebox{0.95\linewidth}{!}{%
\begin{tabular}{cccc} 
\hline
\textbf{Hyperparameter}  & \textbf{Value} & \textbf{Hyperparameter}                                                       & \textbf{Value}  \\ 
\hline
Max Num. epoch           & 8              & Base learning rate                                                            & 0.0003          \\
Early-stopping KL target & 0.024          & Adaptive-LR KL target                                                         & 0.016           \\
Entropy regularization   & 0              & Learning rate schedule                                                        & KL-adaptive        \\
Initial log std.         & -0.4           & log std. decay factor & -0.000367       \\
Policy loss coeff.       & 2              & Value loss coeff.                                                             & 0.5             \\
GAE parameter            & 0.95           & Num. environment                                                              & 1024            \\
Discount factor          & 0.99           & Episode length                                                                & 300             \\
PPO clip range           & 0.3            & Update frequency                                                              & 8               \\
Bound loss coeff.        & 0.02           & Energy loss coeff.                                                            & 8e-5            \\
\bottomrule
\end{tabular}%
}
\end{table}

\subsection{Ablation studies}

\begin{figure}[thbp]
    \centering
    \includegraphics[width=0.9\linewidth]{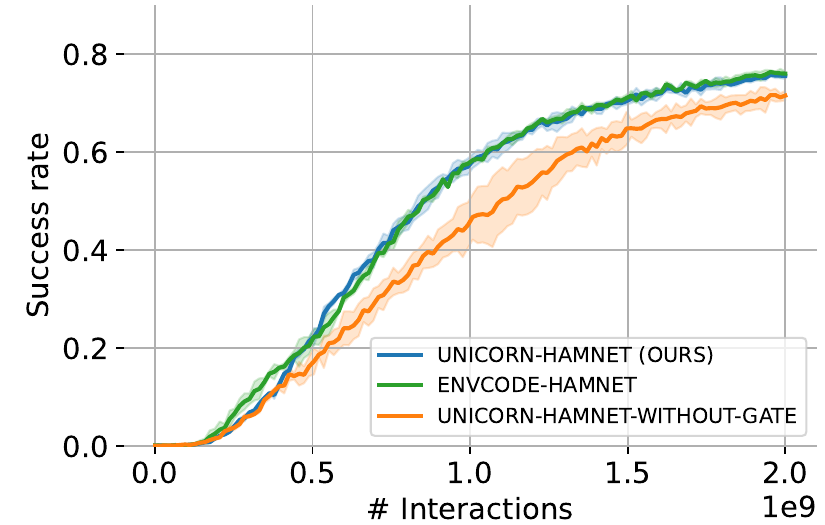}
    \caption{Training progression for our ablation. Plots show the mean (solid) and standard deviations (transparent) for each baseline across three seeds. The interaction steps are reported as the total number of steps aggregated across 1024 parallel environments.}
    \label{fig:abl-result}
    \vspace{-3ex}
\end{figure}

\subsubsection{Effects of the gating mechanism\label{sec:ablation}}
To assess the benefits of the gating mechanism in \oura{}, we compare with \textsc{UniCORN-HAMNet-Without-Gate}, which omits the gating mechanism but is otherwise identical to \textsc{UniCORN}-\oura{}.
Figure~\ref{fig:abl-result} illustrates the training progression for both. While initial trends are similar, \textsc{UniCORN-HAMNet-Without-Gate} trains slower and reaches lower final performance than \ourr{}-\oura{} (71.6\% vs. 75.6\%).
This result highlights the effectiveness of complementing the modules with the gating mechanism to afford additional expressivity.

\subsubsection{Representational quality of \textnormal{\ourr{}} \label{sec:abl-code}}

To evaluate whether \ourr{} sufficiently encodes geometric information about the environment and the object, we compare \ourr{}-\oura{} with \textsc{EnvCode}-\oura{}.
This baseline uses the same architecture as \textsc{UniCORN}-\oura{},
but employs a hand-engineered oracle representation, \textsc{EnvCode}.
This is constructed by concatenating the procedural generation parameters as in Table~\ref{tab:env-code},
yielding a unique 25-dimensional real-valued vector for the scene representation.

\begin{table}[htb]
\centering
\caption{Content of \textsc{EnvCode}.}
\label{tab:env-code}
\resizebox{0.95\linewidth}{!}{%
\begin{tabular}{@{}ccc@{}}
\toprule
\textbf{Parameter} & \textbf{Dimensions} & \textbf{Description} \\ \midrule
ramp position & $\mathbb{R}^{2\times 2}$ & Position of each ramp \\
ramp slope & $\mathbb{R}^{2\times 2}$ & Angle of each ramp \\
plate elevations & $\mathbb{R}^{2\times 3}$ & Height of each base plate \\
wall heights & $\mathbb{R}^4$ & Height of each wall \\
ceiling height & $\mathbb{R}^1$ & Height of the ceiling \\
scene dimension & $\mathbb{R}^3$ & Overall scene dimension \\
scene position & $\mathbb{R}^3$ & Overall scene position \\
\bottomrule
\end{tabular}%
}
\end{table}

As illustrated in Figure~\ref{fig:abl-result}, \textsc{UniCORN-HAMNet} performs on par with \textsc{EnvCode-HAMNet},
achieving a success rate of 75.6\% compared to 76.0\%.
This indicates that \ourr{}, despite operating from sensory observations, provides sufficient information about the environment,
matching the performance of the hand-engineered \emph{oracle} representation
taken directly from the parameters of the underlying procedural generation pipeline.

\subsubsection{Details on the Parameter Scaling Experiment}

In our parameter scaling experiment (see Figure 4), we consider three baselines: \oura{}, \textsc{MLP}, and \textsc{Transformer}.
In \oura{}, we omit the gating to isolate the effects of the modules. This ensures that MLP and \oura{} reduce to the same network architecture when the number of module is one, so that both baselines share the starting point.
As we increase the number of modules, we keep the size of the base network the same in \oura{}, while increasing the width of the hidden layers in the \textsc{MLP} baseline to approximately match the parameter counts, as shown in Table~\ref{tab:mlp-scale}. The \textsc{Transformer} model is considerably larger, with 8 transformer layers, each with the embedding dimension of 512 and 4 heads. All results are aggregated across three different seeds. 

\begin{table}[hbt]
\centering
\caption{MLP baseline configurations scaled to match the parameter count of the \oura{} network for each module count. Network size denotes the dimensions of hidden layers.}
\label{tab:mlp-scale}
\resizebox{0.95\linewidth}{!}{%
\begin{tabular}{ccc}
\toprule
\begin{tabular}[c]{@{}c@{}}\textbf{Corresponding}\\ \textbf{Num. modules}\end{tabular} & \textbf{Network size}        & \textbf{Num. params }\\ \hline
1                                                                    & {[}256, 128, 128, 64{]}  & 0.36 M      \\
2                                                                    & {[}304, 144, 144, 64{]}  & 0.43 M      \\
4                                                                    & {[}512, 256, 256, 128{]} & 0.94 M      \\
8                                                                    & {[}768, 384, 384, 192{]} & 1.76 M     \\
\bottomrule
\end{tabular}%
}
\end{table}

\subsubsection{Details on the Simulation Benchmark\label{sec:r2s_detail}}

\begin{figure*}[tbhp]
    \centering
    \includegraphics[width=1.0\linewidth]{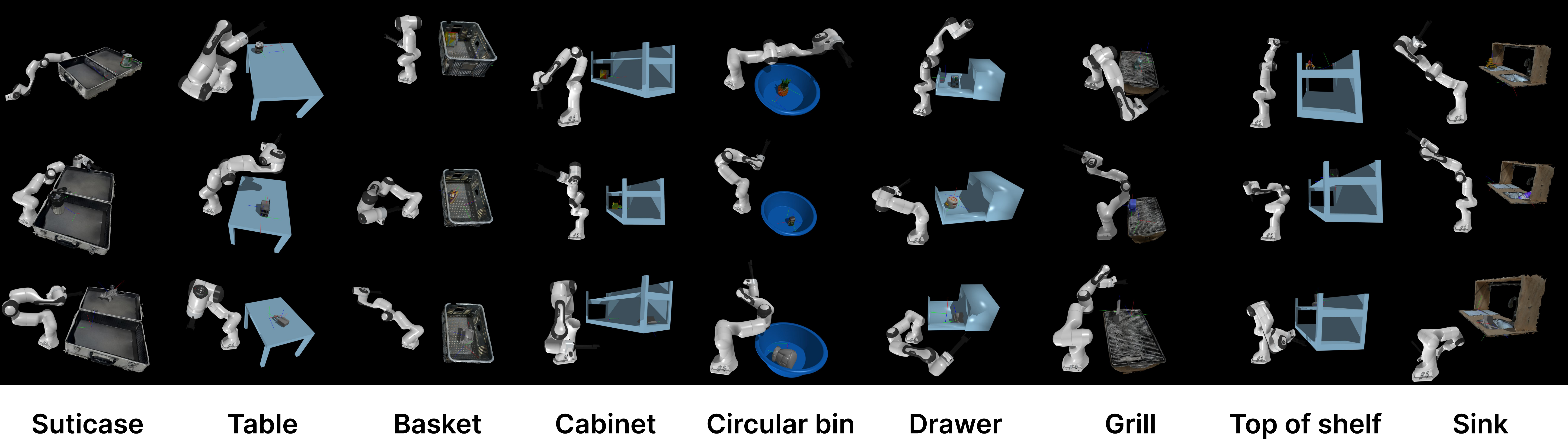}
    \caption{Example configurations for each domain. Each row depicts a different object, and each column shows a different environment.
    }
    \label{fig:r2s-config}
    \vspace{-3ex}
\end{figure*}

\begin{figure}[tbhp]
    \centering
    \includegraphics[width=1.0\linewidth]{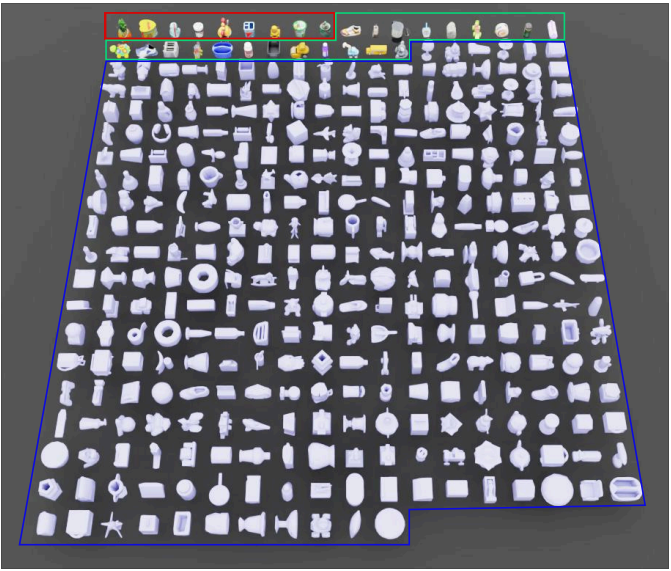}
    \caption{Object meshes used for the benchmark (353 in total). 
    The red box indicates nine custom-scanned objects from the real world; 
    the green box contains 21 objects from GSO;
    and the blue boxes enclose 323 objects from DGN.}
    \label{fig:r2s-objets}
    \vspace{-2ex}
\end{figure}

In our benchmark, we provide nine digital twins of real-world environments paired with a total of 353 objects in our benchmark, shown in Figure~\ref{fig:r2s-objets}: nine custom-scanned objects from the real world (red box), 21 from GSO~\cite{laura2022GSO} (green box), and 323 from DGN~\cite{wang2023dexgraspnet} (blue box).

For the custom-scanned objects, stable poses are collected in the real world using FoundationPose~\cite{wen2024foundationpose}. For the GSO objects, we use Trimesh~\cite{trimesh} to sample their stable orientations. For DGN objects, we use the precomputed stable orientations (see Appendix~\ref{sec:proc-gen-place}).
To determine stable placements for GSO and DGN objects, each object is rotated to one of the pre-sampled stable orientations 
and randomly placed in a predefined region of the environment with a small vertical margin (0.005m). We resolve remaining collisions with FCL~\cite{fcl} to ensure that the object does not penetrate the environment.

From these stable poses, we randomly select pairs of initial and target object poses that maintain sufficient separation in both position and orientation to prevent trivial scenarios. For each domain-object pair, we sample 128 distinct episodes, each with different collision-free robot initializations within the robot’s joint limits, as shown in Figure~\ref{fig:r2s-config}.

\begin{figure}[tbhp]
    \centering
    \includegraphics[width=1.0\linewidth]{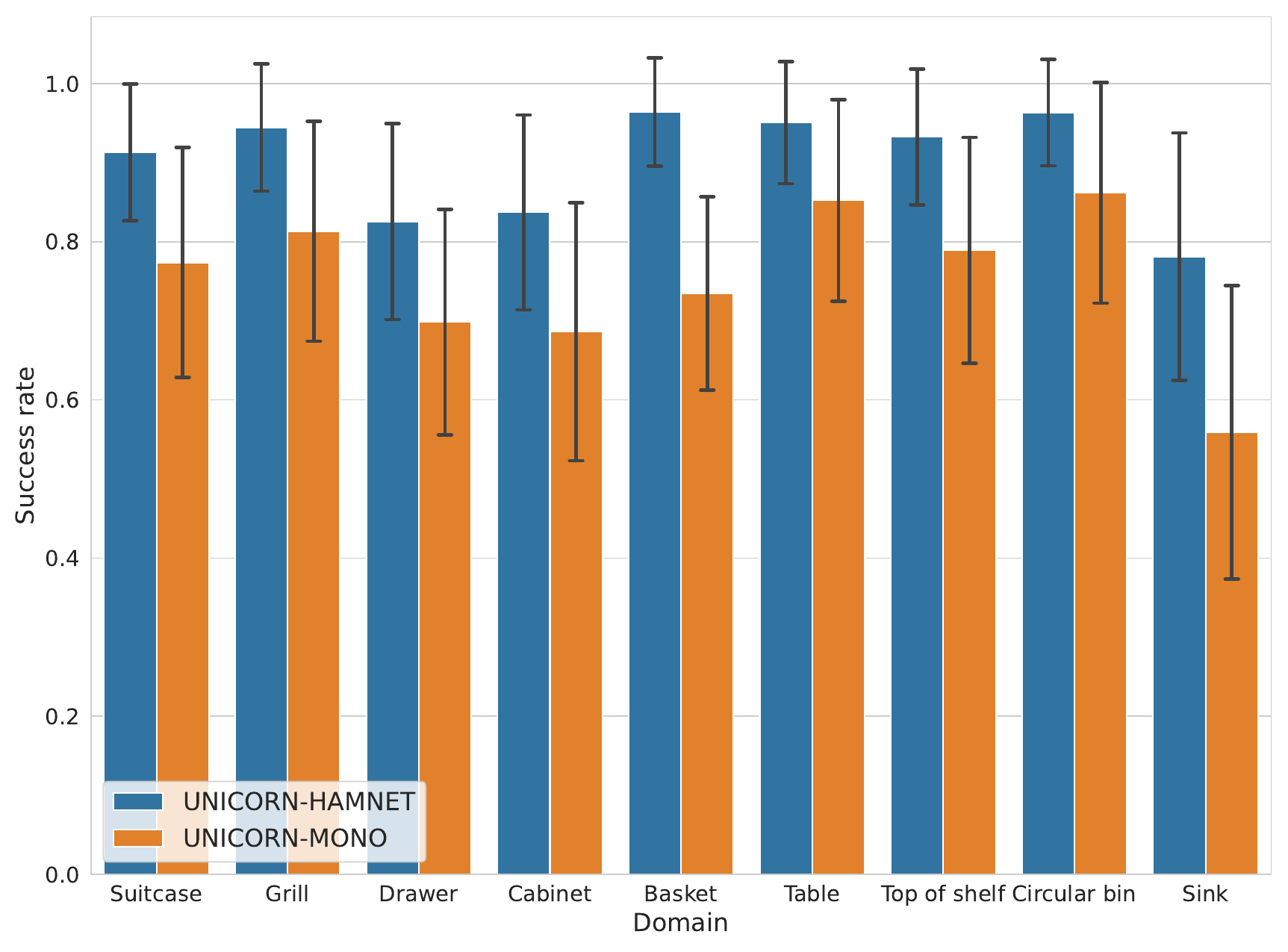}
    \caption{Comparison of success rates of \ourr{}-\oura{} (blue) and \ourr{}\textsc{-MONO} (orange) on our simulated benchmark over 10000 simulation steps.}
    \label{fig:r2s-result}
    \vspace{-3ex}
\end{figure}

To support the evaluation of new algorithms, we establish baseline results from representative monolithic and modular architectures: \textsc{UniCORN-MONO} and \ourr{}-\oura{}, which are illustrated in Figure~\ref{fig:r2s-result}.

\subsection{Details on the Real-World Setup\label{sec:real-world-detail}}

\subsubsection{Pose tracker implementation \label{sec:foundation-pose}}

While the original FoundationPose~\cite{wen2024foundationpose} model only processes one image at a time, we operate with four cameras. To streamline computation, we batchify the model to allow multiple images to be processed at once.
For additional robustness, we also generate multiple pose candidates by adding a small noise (0.02m, 0.15 radians). Each candidate undergoes the refinement procedure as in the original model, then we select the pose with the highest prediction score as input to the policy.

\subsubsection{Failure Modes in the Real World\label{sec:failure-mode}}

We describe our five main failure modes in the real world.
In \emph{torque-limit violation}, the robot aborts due to the robot exceeding the hardware's safety limits. Despite the measures taken in Appendix~\ref{sec:sim2real-transfer}, the sim-to-real gap may still lead to spurious contact in domains with walls, such as the \emph{drawer}, when the robot makes rapid movements to adjust contact sites.
The policy may also get stuck in a \emph{deadlock}, where it indefinitely repeats ineffective maneuvers. For instance, the policy may keep attempting a toppling maneuver for low-friction objects such as \emph{heart-box}, which may not work due to slippage.
In other cases, the robot may accidentally \emph{drop} the object. For example, the object such as the \emph{angled cup} may rapidly bounce or roll off the scene,
and the limited dexterity of our hardware prevents catching such fast-moving objects.
Another failure mode arises when the agent fails to \emph{circumnavigate obstructions}, getting stuck against the environment. This also arises from the sim-to-real gap: while the simulation often allows the robot to move across a shallow barrier by pressing against it, real-world walls cannot be penetrated, which causes the robot to get stuck against the environment.
Lastly, the remaining failures occur from the \emph{perception stack}, where it loses track of the segmentation mask or the object pose. This most frequently occurs when key recognizable textures of the object are occluded by the robot or the environment.

\end{document}